%% file: main.tex
\documentclass[lettersize,journal]{IEEEtran}

\input{preamble}

\usepackage[pagebackref,breaklinks,colorlinks,allcolors=blue]{hyperref}
\usepackage{cleveref}

\begin{document}

\title{G\textsuperscript{3}Splat: Geometrically Consistent Generalizable Gaussian Splatting}

\newcommand{\projectpage}{\href{https://m80hz.github.io/g3splat}{\texttt{m80hz.github.io/g3splat}}\xspace}

\renewcommand{\thefootnote}{\fnsymbol{footnote}}
\setcounter{footnote}{0}

\author{
  Mehdi Hosseinzadeh\textsuperscript{1}\IEEEauthorrefmark{1}\IEEEauthorrefmark{2} \quad
  Shin-Fang Chng\textsuperscript{2}\IEEEauthorrefmark{1} \quad
  Yi Xu\textsuperscript{2} \quad
  Simon Lucey\textsuperscript{1} \quad
  Ian Reid\textsuperscript{1,3} \quad
  Ravi Garg\textsuperscript{1} \\
  \textsuperscript{1} Australian Institute for Machine Learning \quad
  \textsuperscript{2} Goertek Alpha Labs \quad
  \textsuperscript{3} MBZUAI \\
  \projectpage
}

\markboth{IEEE Transactions on Pattern Analysis and Machine Intelligence (TPAMI),~Vol.~xx, No.~xx, 2026}%
{Hosseinzadeh M, Chng SF, \MakeLowercase{\textit{et al.}}: G\textsuperscript{3}Splat: Geometrically Consistent Generalizable Gaussian Splatting}

\twocolumn[{
\renewcommand\twocolumn[1][]{#1}
\maketitle
\input{figs/teaser}
}]

\footnotetext{%
\textsuperscript{*}Equal contribution.\hspace{1.0em}%
\textsuperscript{\textdagger}Corresponding author.%
}
\setcounter{footnote}{0}
\renewcommand{\thefootnote}{\arabic{footnote}}

\input{sec/0_abstract}

\input{sec/1_intro}

\input{sec/2_related_work}
\input{sec/3_method}

\input{sec/4_experiments}

\input{sec/5_conclusion}

{
    \small
    \bibliographystyle{IEEEtran}
    \bibliography{references}
}

\clearpage
\input{sec/appendix}

\end{document}

%% file: preamble.tex
\usepackage[utf8]{inputenc} 
\usepackage[T1]{fontenc}    
\usepackage{url}            
\usepackage{booktabs}       
\usepackage{arydshln}      
\usepackage{amsfonts}       
\usepackage{nicefrac}       
\usepackage{microtype}      
\usepackage{xcolor}         
\usepackage{xspace}
\usepackage{amsmath,amsfonts}
\usepackage{algorithmic}
\usepackage{algorithm}
\usepackage{array}
\usepackage[caption=false,font=normalsize,labelfont=sf,textfont=sf]{subfig}
\usepackage{textcomp}
\usepackage{stfloats}
\usepackage{url}
\usepackage{verbatim}
\usepackage{graphicx}
\usepackage{cite}
\usepackage[normalem]{ulem} 
\usepackage{tabularx}
\usepackage{subcaption}
\usepackage{adjustbox}
\usepackage{multirow}
\usepackage{pifont}
\usepackage{bm}
\usepackage{caption} 
\usepackage{booktabs,adjustbox,siunitx,makecell}
\usepackage[table]{xcolor} 
\usepackage{alphalph}

\newcommand{\best}[1]{\textbf{#1}}

\newcommand{\categorybest}[1]{\underline{#1}} 

\newcommand{\cmark}{\ding{51}}
\newcommand{\xmark}{--}
\newcommand{\src}[1]{\textcolor{darkgray}{#1}}
\newcommand{\NA}{--}
\newcommand{\schemecell}[1]{%
  \footnotesize
  \begin{tabular}[c]{@{}l@{}}#1\end{tabular}%
}

%% file: figs/teaser.tex
\vspace{-3.1em}
\begin{center}
\includegraphics[width=\textwidth]{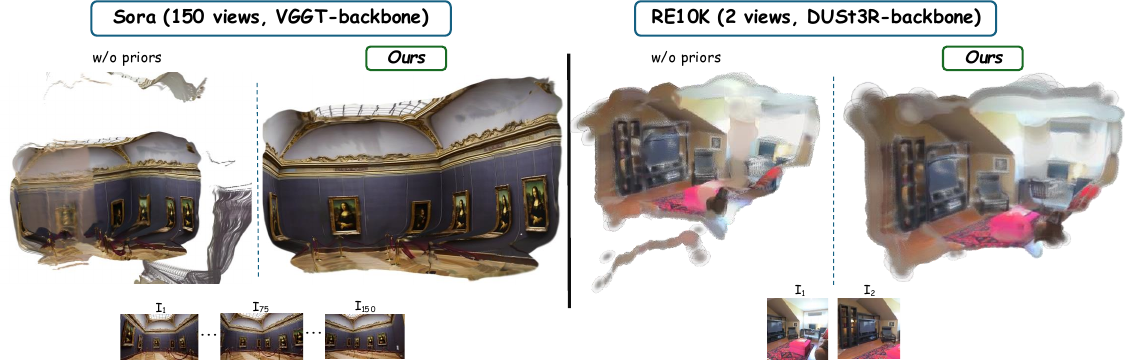}
\captionof{figure}{
\textbf{G\textsuperscript{3}Splat enables geometrically consistent, pose-free generalizable Gaussian splatting across backbones and input regimes.}
We visualize predicted Gaussians \emph{without} and \emph{with} the proposed priors for two representative settings: a VGGT-based~\cite{VGGT} multi-view reconstruction on a Sora-generated sequence (left; 150 input views), and a DUSt3R-based~\cite{dust3r} sparse-view reconstruction on RealEstate10K~\cite{re10k} (right; 2 input views).
Across both cases, the proposed priors improve geometric consistency by reducing floating artifacts, improving cross-view alignment, and producing cleaner Gaussian structure without changing the underlying backbone.
Sora prompt: ``Generate a video inside the Louvre Museum, including the paintings.''
}
\label{fig:teaser1}
\end{center}

%% file: sec/0_abstract.tex
\begin{abstract}

3D Gaussians have become a powerful scene representation for real-time splatting and high-quality novel-view synthesis. This has motivated generalizable splatting -- methods that adapt feed-forward geometry prediction networks to produce per-pixel Gaussians from a set of images. However, most generalizable splatting pipelines are supervised primarily through a view-synthesis loss to predict Gaussian orientation, anisotropic
scale, opacity, and appearance in addition to their locations. 
We show that this learning objective is under-constrained. Models trained with view synthesis alone produce splats whose orientations and scales have no geometric connotation. The result is that, while producing decent view-synthesis performance, nearly all generalizable splatting methods produce geometrically inaccurate and misaligned Gaussians.

We introduce G\textsuperscript{3}Splat, a geometry-consistent generalizable splatting framework that addresses these degeneracies through differentiable geometric priors on the predicted 3D Gaussians, making the learning problem
well-posed. These priors encourage the per-pixel splats to remain on their
viewing rays and to orient themselves in accordance with local surfaces.
Our priors are architecture-agnostic and can be incorporated into any
previously studied geometric backbone for generalizable splatting, as well
as different scene representations. We test G\textsuperscript{3}Splat with both DUSt3R-style and VGGT-style backbones to predict pixel-aligned full-rank
3DGS as well as surfel-like 2DGS.
 
Trained on RE10K, G$^3$Splat produces Gaussian splats with significantly
higher geometric fidelity than baselines, providing state-of-the-art
novel-view depth, mesh reconstruction, and relative pose estimation
performance while preserving novel-view synthesis quality, as evaluated on
datasets such as ACID and ScanNet. Code and pretrained models are released
on our project page.

\end{abstract}


%% file: sec/1_intro.tex
\section{Introduction}
\label{sec:intro}

\IEEEPARstart{G}{aussian} splatting (3DGS)~\cite{3dgs} has rapidly become a dominant representation for 3D structure and appearance modeling from multi-view images. In contrast to explicit depth maps, point clouds, or dense volumetric grids, 3D Gaussians provide a compact differentiable representation that can encode view-dependent appearance and be rendered from arbitrary viewpoints at high speed. These properties have made Gaussian splats an attractive representation not only for per-scene reconstruction, but also for feed-forward scene prediction.

\emph{Generalizable Gaussian splatting} extends 3DGS from per-scene optimization to feed-forward reconstruction of novel views. Instead of fitting Gaussians separately for each scene, recent methods train neural networks that predict 3D Gaussians directly from one or a few input images~\cite{pixelsplat,mvsplat,freesplat,selfsplat,depthsplat,splatt3r,gps-gaussian,noposplat,anysplat,flash3d}. Given these sparse views, the feed-forward networks output a set of Gaussians (means, orientations, scales, and colors), achieving photorealistic novel-view synthesis.

Most existing generalizable splatting frameworks are built by adapting well-established geometry networks originally designed for dense depth~\cite{unimatch,VGGT} or pixelwise 3D point prediction~\cite{dust3r,mast3r,VGGT}. These networks typically use image encoders to process one or multiple images, followed by decoders that predict Gaussian means as per-pixel depth-maps~\cite{unimatch} or 3D points in a canonical frame~\cite{dust3r,mast3r,VGGT} for each frame. Additional decoders are appended to the geometry predictors to predict Gaussian properties such as orientation, scale, opacity, and view-dependent color -- \textit{typically without much foresight}. These networks are usually trained predominantly by minimizing a view-synthesis loss on a few target views, closely following existing self-supervised depth estimators, but with a different image-formation mechanism due to the underlying change in scene representation.

However, this prevalent setup overlooks several key issues inherited from the underlying 3DGS optimization:
\begin{itemize}
  \setlength\itemsep{0.25em}
    \item \textbf{Overparameterization}: 3D Gaussians are overparameterized compared to depth maps or point clouds. Successful estimation of 3D Gaussians typically requires a large number of densely sampled viewpoints. Few-shot 3DGS is known to require priors~\cite{zhu2024fsgs, chung2024depth} even in optimization-based settings; yet generalizable methods typically inherit the 3DGS parameterization without introducing corresponding geometry constraints during training.
    \item \textbf{Geometric ambiguity}: Unlike per-pixel depths or 3D point locations -- which are uniquely defined (up to scale) -- multiple 3D Gaussian configurations can produce equally valid renderings. This ambiguity means that purely image-based loss can let the network converge to geometrically incorrect solutions that still explain the views. 
    \item \textbf{Lack of heuristic}: Successful per-scene Gaussian splatting methods typically rely on non-differentiable heuristics (\textit{i.e.}, splitting, duplication, and pruning of Gaussians). However, existing generalizable methods are trained purely via view-synthesis loss gradients and neglect these heuristics; i.e. all Gaussians remain perpetually alive throughout training.
\end{itemize}

As a result, existing generalizable Gaussian splatting methods often converge to \textit{geometrically degenerate Gaussians}: while the predicted locations (means) remain reasonably accurate — benefiting from well-established depth or point-cloud estimators -- orientations, scales, and opacities are often incorrect. As shown in \Cref{fig:visualize_gaussians}, existing generalizable splatting approaches struggle to learn meaningful opacities, orientations, or scales when trained with view-synthesis loss alone: Gaussian orientations (shown as normal maps) are misaligned with the underlying surface, and 3D Gaussians become unjustifiably elongated or collapsed (shown in their scales). We observe these degeneracies both in splat predictors that anchor Gaussian centers via depth maps~\cite{mvsplat} and in those that use per-pixel 3D point maps aligned to a common reference frame~\cite{noposplat}. Although we highlight these two representative approaches in \Cref{fig:visualize_gaussians}, similar issues appear across \textit{all} existing generalizable splatting methods. We attribute these geometry artifacts to the inherent overparameterization of 3D splats combined with purely photometric supervision: without additional structural priors, self-supervised learning of Gaussian parameter is ill-posed.

Motivated by this analysis, we introduce G\textsuperscript{3}Splat, a geometry-consistent generalizable splatting framework that regularizes Gaussian predictions with two local, differentiable priors. The first prior encourages each Gaussian's orientation to agree with the local surface geometry estimated from neighboring Gaussian means. This prior provides a much direct constraint on the Gaussian orientations, which are left largely unconstrained by the view-synthesis loss. The second prior encourages pixel-aligned 3D Gaussian centers to remain on their corresponding image rays. 
This classic pixel-alignment constraint -- similar to PnP loss -- was not needed when the geometric backbones were trained under full supervision; under view-synthesis-based supervision, however, it become crucial. 
Together, these priors provide direct geometric guidance for the Gaussian parameters most weakly constrained by view synthesis alone.

The proposed priors are deliberately kept architecture-agnostic: they operate directly on the predicted Gaussian centers and covariances, and can therefore be used with different feed-forward geometry backbones~\cite{unimatch,dust3r,mast3r}, and different splat parameterizations~\cite{2DGS,3dgs}. 
We study the advantages of the proposed priors with two different geometric backbone architectures: the two-view, canonical point-map prediction network of DUSt3R~\cite{dust3r}, and the more versatile, better-performing VGGT backbone~\cite{VGGT} that can take multiple images as input. 
The simpler DUSt3R adaptation provides a controlled setup to analyze and isolate the need for, and the efficacy of, the proposed priors,\footnote{VGGT provides relative camera poses, per-frame canonical point clouds, depth maps, and point tracks. Many of these estimates are redundant and not mutually aligned, offering several alternative ways to combine them for generalizable splatting. We leave exploring these design choices to future work and ``splatify'' VGGT analogously to our DUSt3R-based framework.} and is used for a detailed study of the presented approach. 
In particular, we evaluate the improvements in multi-view Gaussian alignment, rendered depth, and mesh reconstruction, for both surfel-like 2D Gaussians~\cite{2DGS} and standard 3D Gaussians\cite{3dgs}. 
We additionally use relative pose estimation as a stress test of geometric consistency: if the predicted splat centers are coherent across views, a classical PnP solver should recover accurate relative poses without relying on photometric pose refinement. 
We demonstrate that these gains transfer seamlessly to VGGT adaptations as well. Thanks to VGGT's capability to operate on hundreds of frames, we show that our approach yields geometrically accurate splat predictions superior to alternatives such as AnySplat~\cite{anysplat}, which \textit{splatifies} VGGT.
Importantly, the proposed priors preserve -- and in most cases slightly improve -- novel-view synthesis quality, indicating that the gains in geometry do not come at the expense of image rendering.

\input{figs/gaussian_params_comparisons}

To the best of our knowledge, this is the first work in generalizable Gaussian splatting to systematically analyze and address the geometric degeneracies of predicted Gaussian orientations and anisotropic scales. 
We identify the under-constrained degrees of freedom in self-supervised splat prediction and introduce simple differentiable priors that can be integrated into existing geometry predictors. 
This makes the approach applicable to depth- and point-map-based splatting pipelines, as well as broader multi-view geometry backbones.

The main contributions of this paper are:
\begin{itemize}
  \setlength\itemsep{0.25em}
  \item We identify and analyze the geometric degeneracies of generalizable Gaussian splatting trained with view-synthesis supervision alone.
  \item We propose two simple, differentiable, and architecture-agnostic geometric priors -- orientation consistency and pixel alignment -- that supervise the splat parameters left unconstrained by view synthesis.  
  \item We show that the priors improve both full-rank 3DGS and surfel-like 2DGS, and transfer from DUSt3R-style to VGGT-style backbones.  
  \item We achieve state-of-the-art rendered depth, mesh reconstruction, and relative pose estimation among (pose-free) generalizable splatting methods, with strong zero-shot generalization to ScanNet and ACID.
  \item We show that these geometric gains preserve, and often slightly improve, novel-view synthesis quality.
\end{itemize}

Overall, G$^3$Splat provides a simple recipe for learning Gaussian splats whose parameters are not merely useful for rendering, but are geometrically consistent enough to support 3D reconstruction, pose estimation, and downstream geometric reasoning.

%% file: figs/gaussian_params_comparisons.tex
\begin{figure*}[ht!]
  \centering
  \includegraphics[width=\textwidth]{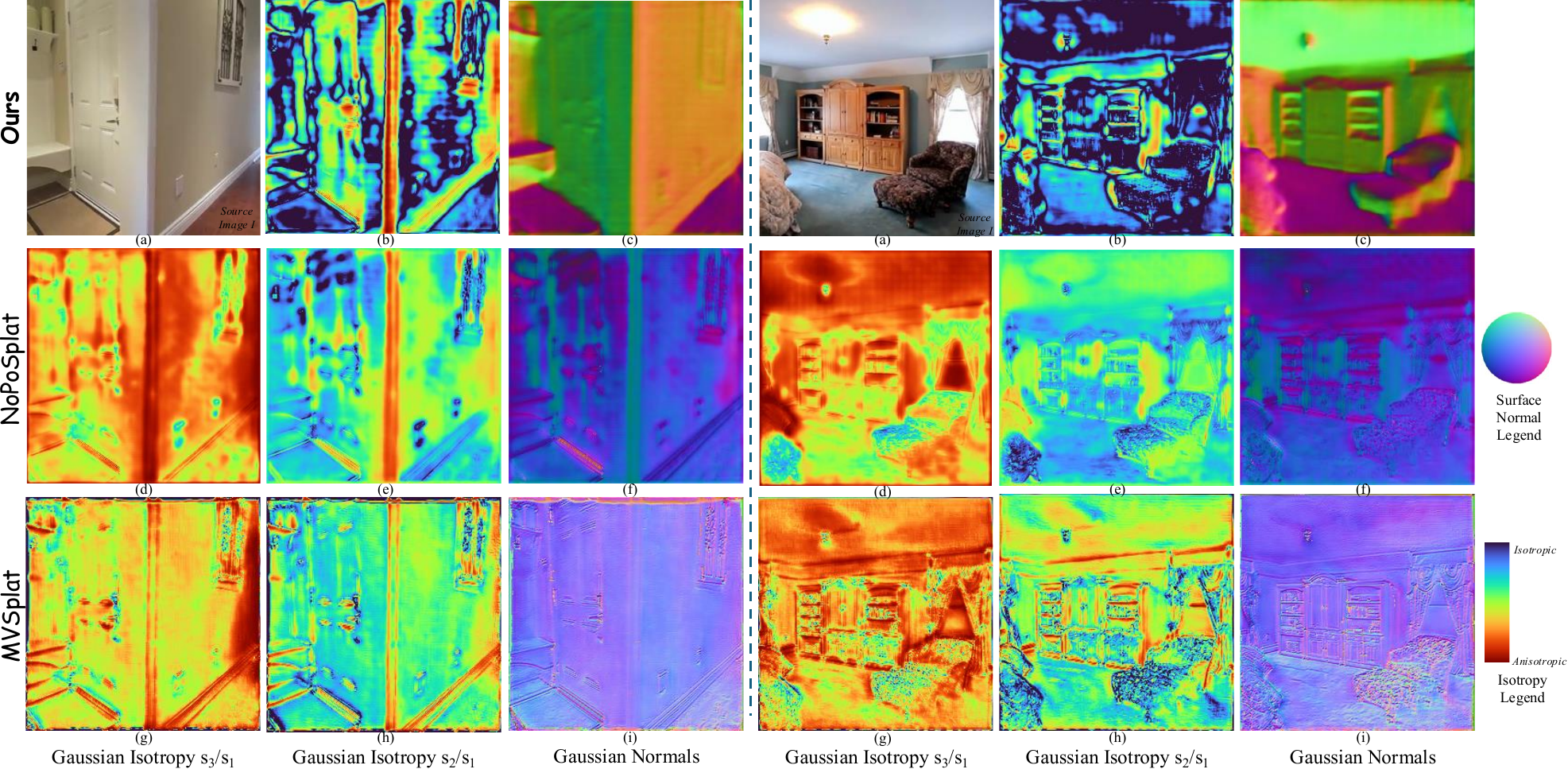}
\caption{\textbf{Qualitative comparison of predicted Gaussian parameters.}
For visualization, we denote by $(s_1,s_2,s_3)$ the \emph{sorted} eigen-scales of each Gaussian covariance in \Cref{eq:covariance}, such that $s_1 \ge s_2 \ge s_3$; the smallest scale $s_3$ characterizes uncertainty along the surface normal direction.
\textbf{Row 1 (ours)} shows: (a) the source image to which Gaussians are aligned, (b) skewness of the estimated Gaussians within their defining plane, and (c) predicted Gaussian orientations visualized as surface-normal maps.
\textbf{Rows 2 and 3} show results for NoPoSplat~\cite{noposplat} and MVSplat~\cite{mvsplat}, respectively: (d/g) Gaussians' elongation perpendicular to the dominant plane defined by it, (e/h) Gaussians' skewness within the dominant plane, and (f/i) normals to the dominant plane.
Existing methods yield Gaussian orientations with limited geometric meaning: 
MVSplat Gaussians (i) align mostly fronto-parallel to the source image plane, and NoPoSplat Gaussians orientations (f) strongly depend on texture, spanning a few dominant directions inconsistent with scene geometry. Our method produces plausible, near-Manhattan structured surface orientations. Baseline Gaussians exhibit significant elongation perpendicular to their dominant surfaces (visible as non-red colors in d/g). Notably, our Gaussians remain relatively circular (blue color in b) on planar, textureless surfaces and become skewed ellipses (red color in b) near sharp geometric edges such as shelves or wall corners.}
\label{fig:visualize_gaussians}
\end{figure*}

%% file: sec/2_related_work.tex
\section{Related Work}
\label{sec:related}

Owing to the state-of-the-art real-time view synthesis performance of 3D Gaussian splatting \cite{3dgs}, significant effort has been put into improving 3DGS for scenarios such as few-view reconstruction \cite{zhu2024fsgs, chung2024depth, li2024dngaussian, binocular3dgs, xiong2023sparsegs, s2gaussian}, dynamically moving objects \cite{wu2024_4dgs, yang2023real, yang_2024, li_2024}, surface extraction \cite{sugar, 2DGS}, and incorporating object semantics into 3D reconstructions \cite{li2024sgs}. Real-time simultaneous localization and mapping approaches have also adapted Gaussian splats as an inherent scene representation \cite{monogs, splatam}. Additionally, Gaussian splats have been used for generating geometrically consistent images and video sequences \cite{cut3r, wu2024_4dgs}. 
Most related to our priors are per-scene methods that explicitly regularize splat geometry: 2DGS~\cite{2DGS} enforces consistency between rendered normals and rendered depth, while SuGaR~\cite{sugar} encourages Gaussians to align with, and remain flat on, the underlying surface. These regularizers, however, operate within per-scene optimization. We bring this geometric-regularization perspective to feed-forward, self-supervised splat prediction, where the constraints must act directly on network outputs and remain stable under a purely photometric training signal (see \Cref{sec:supp_rendered_normal} for a comparison with the rendered normal--depth consistency loss of~\cite{2DGS}).

The deep learning revolution of the last decade has significantly influenced geometric inference from one or more images. Earlier works focused on training neural networks to map a single image to depth maps obtained from range sensors \cite{eigen2014depth, laina2016deeper, fu2018deep, lee2019bts, dpt}.  
Multi-view extensions for these supervised learning algorithms are well explored as well \cite{chang2018pyramid, huang2018deepmvs, mvsnet, casmvsnet, patchmatchnet, ummenhofer2017demon}. 
More recently, methods have explored reconstructing registered sets of per-pixel point clouds from multiple images, providing state-of-the-art relative pose and scene structure \cite{dust3r, mast3r}. 

Additionally, it has been demonstrated that these feed-forward geometry predictors can be trained without depth sensors in a self-supervised manner by minimizing view synthesis losses \cite{garg2016, godard2017, zhou2017, zhan2018, godard2019}. Structure prediction from single or few images has also been utilized as an optimization-free building block in high-fidelity tracking and mapping systems \cite{niceslam, nicerslam}. 
Generalizable Gaussian Splatting methods have evolved recently to learn neural networks that predict 3D Gaussians explaining a scene directly from a few images. We broadly categorize these methods into two categories:

\textbf{Pose-Dependent Generalizable 3DGS}: Several works assume input images come with known or precomputed poses (e.g., via SfM) and focus on designing architectures to infer 3D Gaussians from these posed views \cite{pixelsplat, mvsplat, efreesplat, freesplat, depthsplat, colmap-free3DGS, hisplat, gps-gaussian}. A prominent example is pixelSplat~\cite{pixelsplat}, which introduced a two-view feed-forward network that utilizes epipolar cross attention transformer architecture to fuse multi-view information and predict per-pixel depth distribution for input images. These distributions are sampled to create a set of 3D Gaussian centers along the viewing rays. MVSplat~\cite{mvsplat} uses cost volume based fusion of multi-view information, adapting the Unimatch~\cite{unimatch} architecture to regress for depth instead. Both methods use additional decoder heads to estimate the remaining 3D Gaussian parameters. 

\textbf{Pose-Free Generalizable 3DGS}: An emerging frontier involves dispensing with known camera poses—allowing the network to infer scene geometry and camera registration jointly from images alone \cite{splatt3r, selfsplat, noposplat, anysplat}. Early efforts in this direction often build upon learned stereo matching. For example, \cite{splatt3r} tackles uncalibrated stereo pairs by extending a foundation model (MASt3R \cite{mast3r}) that predicts dense point clouds from two images. It then outputs 3D Gaussians directly in a canonical frame, augmenting each point in the MASt3R reconstruction with color and covariance attributes. This process is supervised using the geometry of the 3D point cloud and followed by a novel-view synthesis stage to fine-tune appearance. NoPoSplat \cite{noposplat} adopts a more self-supervised, multi-view approach by anchoring one view's coordinate system as canonical and training a network to predict all Gaussians directly in that space, using only a photometric loss for training. 
More recently, AnySplat~\cite{anysplat} splatifies the multi-view VGGT backbone~\cite{VGGT}, but relies on pseudo-depth supervision distilled from the backbone together with additional training data. In contrast, we retain purely self-supervised view-synthesis training and instead make the learning problem well-posed through explicit, differentiable geometric priors.

To the best of our knowledge, all aforementioned generalizable splatting methods struggle to learn geometrically faithful orientations and scales for 3D Gaussians. The proposed approach addresses this issue in generalizable splatting through explicit geometric priors.

%% file: sec/3_method.tex
\section{Method}
\label{sec:method}
In this section, we present our generalizable Gaussian splatting formulation and the regularization losses used to make self-supervised splat prediction geometrically better posed. The formulation is independent of a particular backbone; architectural instantiations are described in the supplementary material.

\noindent\textbf{Problem Definition.} Given a set of sparse images $\mathcal{I} = \{\bm{I}_t \in \mathbb{R}^{H \times W \times 3} \}_{t=1}^T$ (also referred to as context images in~\cite{pixelsplat,splatt3r, mvsplat, noposplat, depthsplat,selfsplat}), each with known camera intrinsics that form the set $ \mathcal{K} = \{\bm{K}_t\in \mathbb{R}^{3 \times 3}\}_{t=1}^T$ capturing a \textit{rigid} scene, we learn a feedforward neural network $f_{\Theta}$ that maps these images and intrinsics $(\mathcal{I}, \mathcal{K})$ to a set of \textit{pixel-aligned} Gaussians as: 
\begin{equation}
    f_{\bm \Theta}( \mathcal{I}, \mathcal{K}) = \left\{
    \mathcal{G}_t^j :=\left(\boldsymbol{\mu}_t^j, \alpha_t^j, \boldsymbol{ q}_t^j, \boldsymbol{s}_t^j, \boldsymbol{c}_t^j\right)\right\}_{t=1:T}^ {j=1:H \times W},
    \label{eq:mapping}
\end{equation}
\noindent where $\mathcal{G}_t^j$ is the 3D Gaussian defined in the 3D space corresponding to a pixel $j$ in image $t$. Each $\mathcal{G}_t^j$ is characterized by its \textit{center} $\boldsymbol{\mu} \in \mathbb{R}^3$; \textit{orientation} represented by a unit quaternion vector $\boldsymbol{q} \in$ $\mathbb{R}^4$; three \textit{scale} parameters $\boldsymbol{s} \in \mathbb{R}^3$ defining the elongation of the 3D Gaussians; \textit{opacity} $\alpha \in \mathbb{R}$; and \textit{color} encoded as spherical harmonics $\boldsymbol{c} \in \mathbb{R}^d$. 
In addition to the prevalent 3DGS representation adopted by generalizable Gaussian splatting frameworks~\cite{pixelsplat,mvsplat,depthsplat,noposplat}, we also explore the 2DGS representation introduced by~\cite{2DGS}, which models the scene with 2D surfels instead of standard 3D Gaussians. 
We present extensive evaluations and ablations quantifying the impact of this representation on generalizable Gaussian splatting in \Cref{sec:experiments} and in the supplementary material.

Note that both Gaussian centers $\boldsymbol{\mu}_t^j$ and orientations $\boldsymbol{q}_t^j$ are defined in the camera coordinate frame of the \textit{first} image $\bm{I}_1$. 
Given these $H\times W \times T$ Gaussian predictions, we render \textit{novel views} of the scene $ \{\hat{\mathbf{I}}_f \in \mathbb{R}^{H\times W \times 3} \}_{f=1}^F$ from $F$ different viewpoints defined by the projection matrix $P_f = (\mathbf{R}_f,\mathbf{T}_f) \in \text{SE(3)}$ to be matched with the observed image $\mathbf{I}_f$ during training. 

We minimize a view-synthesis loss~\cite{noposplat,mvsplat,pixelsplat} between the ground-truth image $\mathbf{I}_f$ and the rendered image $\hat{\mathbf{I}}_f$:
\begin{equation}
    \mathcal{L}_{\text{synthesis}}
    = \sum_{f=1}^F \Bigl[
        \mathcal{L}_{\text{rgb}}(\mathbf{I}_f , \hat{\mathbf{I}}_f)
        + \mathcal{L}_{\text{lpips}}(\mathbf{I}_f , \hat{\mathbf{I}}_f)
      \Bigr].
    \label{eq:synthesis_loss}
\end{equation}
For a pixel $(u,v)$ in view $f$, the rendered color $\hat{\mathbf{I}}_f(u,v)$ is obtained by alpha-blending $K$ depth-sorted projected Gaussians $\mathcal{G}'_k$:
\begin{align}
    \hat{\mathbf{I}}_f(u,v)
    &= \sum_{k=1}^K \boldsymbol{c}_k \, w_k(u,v),
    \label{eq:synthesis_render} \\
    w_k(u,v)
    &= T_k(u,v)\,\alpha_k\,\mathcal{G}'_k(u,v), \\
    T_k(u,v)
    &= \prod_{i<k} \bigl(1 - \alpha_i\,\mathcal{G}'_i(u,v)\bigr),
\end{align}
where $\boldsymbol{c}_k$ and $\alpha_k$ denote the color and opacity of Gaussian $k$, and $\mathcal{G}'_k(u,v)$ is its 2D footprint in the image plane of $\mathbf{I}_f$ (see supplementary material for details).

As shown in \Cref{sec:experiments}, a view-synthesis loss alone is insufficient for learning geometrically meaningful Gaussians. In this work, we propose to minimize two additional regularization terms: (i) an orientation-consistency term $\mathcal{L}_{\text{orient}}$ to align the orientations of the Gaussians with the dominant local surface normal; (ii) a pixel-alignment loss $\mathcal{L}_{\text{align}}$ to ensure that the estimated Gaussians remain aligned with the pixels of the provided images. Combining these two regularizers with the view-synthesis loss, we define our training objective function $\mathcal{L}_{\text{total}}$ as
\begin{equation}
    \mathcal{L}_{\text{total}} = \mathcal{L}_{\text{synthesis}} + \lambda_{o} \mathcal{L}_{\text{orient}}  + \lambda_{a}  \mathcal{L}_{\text{align}},
    \label{eq:total_loss}
\end{equation}
where $\lambda_{o}$ and $\lambda_{a}$ are weighting factors balancing the influence of each regularization (see \Cref{sec:supp_implementation_details}). We discuss the motivation, formulation, and effect of each regularizer in the following sections.

\subsection{Learning Gaussian Orientations}

As discussed in \Cref{sec:intro}, existing pose-free and pose-aware generalizable splatting approaches often learn Gaussian orientations with weak geometric meaning. To give these orientations a geometric interpretation, we align them with the dominant surface normals of the underlying scene.

In both the 3DGS and 2DGS parameterizations, each Gaussian $\mathcal{G}_t^j$ is associated with a covariance matrix
\begin{align}
    \boldsymbol{\Sigma}_t^j 
    = \mathbf{R}(\boldsymbol{q}_t^j)\,
      \mathrm{diag}\big([s_t^{j,1}, s_t^{j,2}, s_t^{j,3}]^\top\big)\,
      \mathbf{R}(\boldsymbol{q}_t^j)^\top,
      \label{eq:covariance}
\end{align}
where $\mathbf{R}(\boldsymbol{q}_t^j) \in \mathrm{SO}(3)$ is the rotation induced by the quaternion $\boldsymbol{q}_t^j$ and $s_t^{j,1},s_t^{j,2},s_t^{j,3}\ge 0$ denote the axis-aligned scales in the local Gaussian frame.
We define the \emph{Gaussian normal} $\boldsymbol{N}_t^j$ as the unit eigenvector of $\boldsymbol{\Sigma}_t^j$ corresponding to its smallest eigenvalue (equivalently, the column of $\mathbf{R}(\boldsymbol{q}_t^j)$ associated with $\arg\min_k s_t^{j,k}$).

In the 2DGS variant~\cite{2DGS}, each Gaussian is surfel-like and is parameterized by only two in-plane scales. For notational consistency with \Cref{eq:covariance}, we represent this by appending a third scale fixed to zero, i.e., we set $s_t^{j,3}=0$ in the diagonal of $\boldsymbol{\Sigma}_t^j$. This yields a rank-deficient covariance whose null-space direction defines $\boldsymbol{N}_t^j$ as in~\cite{2DGS} (and $\arg\min_k s_t^{j,k}$ is attained by the appended zero scale).
Although this 2DGS specialization reduces over-parameterization, the view-synthesis loss in \cref{eq:synthesis_loss} still provides only weak supervision for learning orientations, for both 3DGS and 2DGS. 
A natural alternative is the rendered normal--depth consistency regularizer from~\cite{2DGS}; however, naively deploying this rasterization-based regularizer in our generalizable setting does not yield stable training. We analyze this behavior and discuss remedies in the supplementary material.

\noindent\textbf{Orientation supervision from local geometry.}
Instead, we directly supervise orientations using local surface normals estimated from neighboring Gaussian means. Assuming each Gaussian $\mathcal{G}_t^j$ is aligned with pixel $j=(u,v)$ in frame $t$, we define central differences
\begingroup
\begin{align}
\Delta_x \boldsymbol{\mu}_t^{(u,v)}
&= \boldsymbol{\mu}_t^{(u+1,v)} - \boldsymbol{\mu}_t^{(u-1,v)}, \nonumber\\
\Delta_y \boldsymbol{\mu}_t^{(u,v)}
&= \boldsymbol{\mu}_t^{(u,v+1)} - \boldsymbol{\mu}_t^{(u,v-1)}, 
\label{eq:central_differences}
\end{align}
\endgroup
and compute a \emph{local surface normal} from the corresponding 3D points as
\begingroup
\begin{align}
    \hat{\boldsymbol{N}}_t^{j}
    &= \Big\|
        \Delta_y \boldsymbol{\mu}_t^{(u,v)}
        \times
        \Delta_x \boldsymbol{\mu}_t^{(u,v)}
      \Big\|_{*},
      \label{eq:normal_from_means}
\end{align}
\endgroup
where $\|\cdot\|_{*}$ denotes vector normalization and $\boldsymbol{\mu}_t^{(u,v)}$ is the 3D mean of the Gaussian aligned with pixel $(u,v)$.

\noindent\textbf{Edge-aware orientation loss.}
Normals derived from local finite differences are least reliable near depth discontinuities. To reduce the influence of such pixels, we use an edge-aware weight based on the local 3D variation:
\begingroup
\begin{align}
d_t^{j}
&= \big\|\Delta_x \boldsymbol{\mu}_t^{(u,v)}\big\|_2
 + \big\|\Delta_y \boldsymbol{\mu}_t^{(u,v)}\big\|_2, 
 \nonumber \\
\eta
&= \mathrm{Quantile}_{q}\!\left(\{d_t^{j}\}\right),
\nonumber \\
w_t^{j}
&= w_0 \exp\!\Big(-\kappa\, d_t^{j} / (\eta+\epsilon)\Big), 
\label{eq:edge_aware_weights}
\end{align}
\endgroup
with fixed constants $(w_0,\kappa)$, $q$-quantile normalization, and a small $\epsilon$ for stability (see \Cref{sec:supp_implementation_details}).
We then encourage the predicted Gaussian normal $\boldsymbol{N}_t^j$ (smallest-eigenvalue direction for 3DGS, null-space direction for 2DGS) to agree with $\hat{\boldsymbol{N}}_t^j$ using a Huber penalty in cosine space:
\begingroup
\begin{align}
    \mathcal{L}_{\text{orient}}
    = \frac{1}{|\Omega|}
      \sum_{(t,j)\in\Omega}
      w_t^{j}\,
      \mathcal{H}_{\delta}\!\Big( 1 - \langle \boldsymbol{N}_t^{j}, \hat{\boldsymbol{N}}_t^{j} \rangle \Big),
    \label{eq:orient_loss}
\end{align}
\endgroup
where $\langle\cdot,\cdot\rangle$ denotes the dot product, $\mathcal{H}_{\delta}(\cdot)$ is the Huber loss with threshold $\delta$, and $\Omega$ denotes the set of valid interior pixels over all frames (excluding a one-pixel boundary where central differences are undefined), i.e., $|\Omega| = T(H-2)(W-2)$.

Unlike rasterization-based priors, $\mathcal{L}_{\text{orient}}$ depends only on the predicted Gaussian means and covariances, providing direct supervision of orientation given $\boldsymbol{\mu}_t^j$. This formulation is representation-agnostic and applies equally to 3DGS and 2DGS. Conceptually, it mirrors supervised surface-normal training~\cite{eigen2015predicting}, where reference normals are derived from depth via \Cref{eq:normal_from_means}.
In our experiments, this formulation is more stable than alternative orientation regularizers and can also be used in depth-supervised training of generalizable Gaussian splats.

\noindent\textbf{Scale regularization (3DGS only).}
In our 3DGS variant, we additionally apply an anisotropy bias on the Gaussian scales to discourage near-isotropic covariances. Concretely, we add a penalty on the minimum per-Gaussian scale,
\begingroup
\begin{align}
\mathcal{L}_{\text{flat}}
&=
\frac{1}{\mathcal{P}}
\sum_{t,u,v}\min\!\bigl(s_t^{(u,v),1},\,s_t^{(u,v),2},\,s_t^{(u,v),3}\bigr),
\nonumber \\
\mathcal{L}_{\text{total}}
&\leftarrow
\mathcal{L}_{\text{total}} + \lambda_{\text{flat}}\,\mathcal{L}_{\text{flat}} .
\end{align}
\endgroup
where the scales are those used in \Cref{eq:covariance}, $\lambda_{\text{flat}}$ is the weight, and $\mathcal{P}=TWH$ is the total number of pixel-aligned Gaussians in the scene.
Unless stated otherwise, for 3DGS we apply this term together with the orientation loss (i.e., whenever $\mathcal{L}_{\text{orient}}$ is enabled).
This term is disabled for 2DGS, where $s_t^{(u,v),3}=0$ by construction.

\subsection{Pixel-aligned Gaussians}

Pose-aware generalizable splatting methods~\cite{pixelsplat} often adapt depth-prediction networks and, by construction, constrain every Gaussian to lie on its corresponding viewing ray. Pose-free variants~\cite{noposplat} drop the camera-pose assumption by directly estimating Gaussian locations in a canonical space using a DPT decoder. While this removes the need to warp Gaussians with known cameras, the resulting parameterization renders structure estimation ill-posed, especially in the self-supervised regime. In contrast to depth-supervised frameworks such as DUSt3R~\cite{dust3r}, which learn an implicit structural prior by enforcing the reconstructed 3D point cloud to project onto a regular image grid, the pure view-synthesis loss in \cref{eq:synthesis_loss} does not impose such a constraint. Gaussians can therefore move freely into geometrically degenerate configurations, degrading both structure and relative pose estimation, while still explaining the input appearance. 

We therefore explicitly align each Gaussian with its pixel's viewing ray. 
Concretely, for each pixel $(u,v)$ in frame $t$, the Gaussian center $\boldsymbol{\mu}_t^{(u,v)}$ should reproject to $(u,v)$ under the corresponding camera intrinsics $\mathbf{K}_t$ and extrinsics $(\mathbf{R}_t,\mathbf{T}_t)$. We enforce this constraint with the alignment loss
\begingroup
\begin{align}
&\mathcal{L}_{\text{align}}
=
\frac{1}{\sum_{t,u,v}\mathcal{M}_t^{(u,v)}}
\sum_{t=1}^{T}\sum_{u=1}^{W}\sum_{v=1}^{H}
\mathcal{M}_t^{(u,v)}\,\ell_{\text{align}}(u,v,t),
\label{eq:align_loss}\\
&\ell_{\text{align}}(u,v,t)
=
\big\|
\begin{bmatrix}u&v\end{bmatrix}^{\!\top}-\Pi\!\big(\mathbf{K}_t[\mathbf{R}_t~|~\mathbf{T}_t]\tilde{\boldsymbol{\mu}}_t^{(u,v)}\big)
\big\|_2^2,
\label{eq:align_term}
\end{align}
\endgroup
where $\tilde{\boldsymbol{\mu}}_t^{(u,v)}=\begin{bmatrix}(\boldsymbol{\mu}_t^{(u,v)})^\top&1\end{bmatrix}^{\!\top}$,
$\mathcal{M}_t^{(u,v)}{=}1$ if the projected mean lies within the image bounds and has positive depth (and $0$ otherwise), and
$\Pi([X,Y,Z]^\top)=[X/Z,\;Y/Z]^\top$ denotes the perspective projection.

As shown in \Cref{sec:experiments}, $\mathcal{L}_{\text{align}}$ plays an important role in PnP-based relative pose estimation
(\Cref{tab:pose_comparison_all,tab:pose_comparison_details}) as well as accurate depth and structure estimation
(\Cref{tab:scannet_depth_novel_source}).

%% file: sec/4_experiments.tex
\section{Experiments}
\label{sec:experiments}

\begin{figure*}[!t]
\centering
\setlength{\tabcolsep}{1pt}
\begin{tabular*}{\textwidth}{@{} c c c c c c c@{}}
\textbf{Inputs} & \textbf{pixelSplat} & \textbf{MVSplat} & \textbf{DepthSplat} & \textbf{NoPoSplat} & \textbf{Ours} &\textbf{GT RGB/Depth}\\ 
    \begin{minipage}[b]{0.075\textwidth}
        \includegraphics[width=\textwidth]{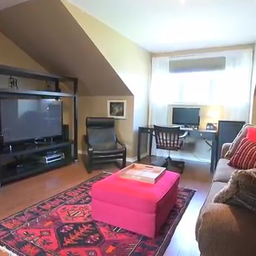}\\[3pt]
        \includegraphics[width=\textwidth]{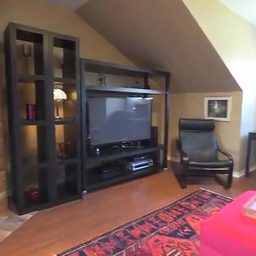}
    \end{minipage} 
    &
    \includegraphics[width=0.15\textwidth]{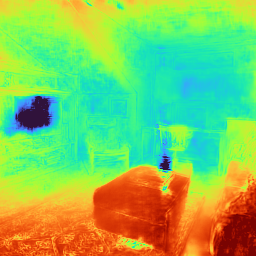}
    &
    \includegraphics[width=0.15\textwidth]{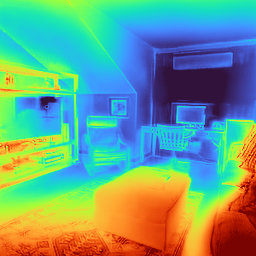}
    &
    \includegraphics[width=0.15\textwidth]{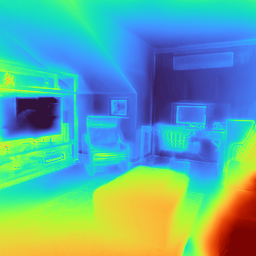}
    &
    \includegraphics[width=0.15\textwidth]{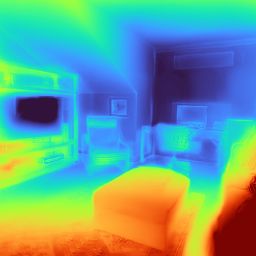}
    &
    \includegraphics[width=0.15\textwidth]{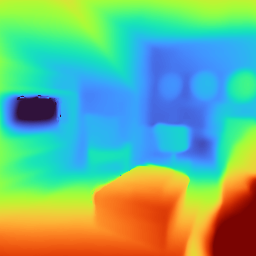}
    &
    \includegraphics[width=0.15\textwidth]{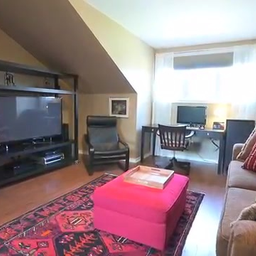}
    \\
    \begin{minipage}[b]{0.075\textwidth}
        \includegraphics[width=\textwidth]{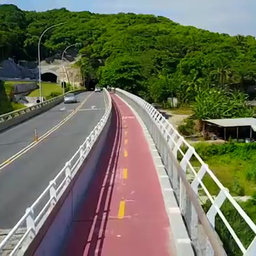}\\[3pt]
        \includegraphics[width=\textwidth]{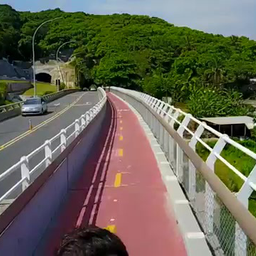}
    \end{minipage}
    &
    \includegraphics[width=0.15\textwidth]{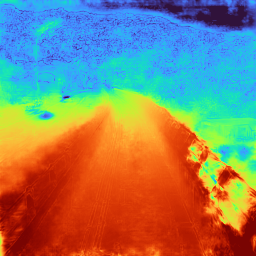}
    &
    \includegraphics[width=0.15\textwidth]{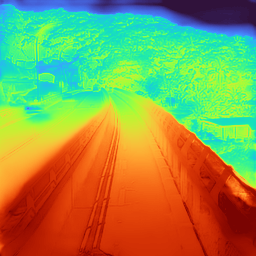}
    &
    \includegraphics[width=0.15\textwidth]{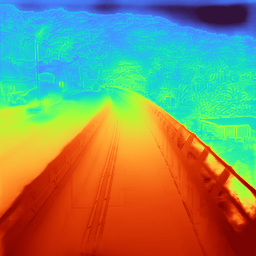}
    &
    \includegraphics[width=0.15\textwidth]{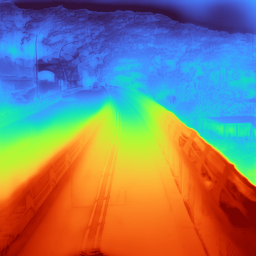}
    &
    \includegraphics[width=0.15\textwidth]{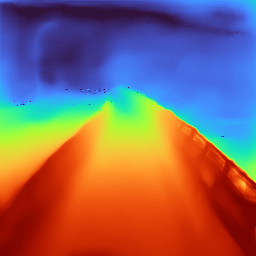}
    &
    \includegraphics[width=0.15\textwidth]{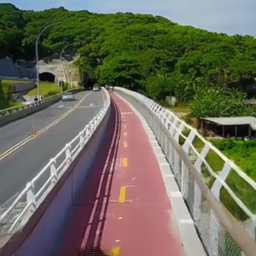}
    \\
    \begin{minipage}[b]{0.075\textwidth}
        \includegraphics[width=\textwidth]{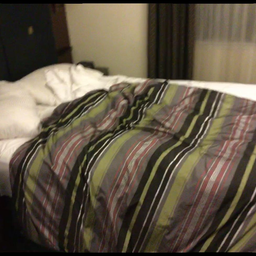}\\[3pt]
        \includegraphics[width=\textwidth]{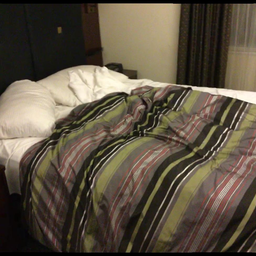}
    \end{minipage}
    &
    \includegraphics[width=0.15\textwidth]{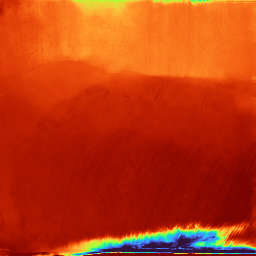}
    &
    \includegraphics[width=0.15\textwidth]{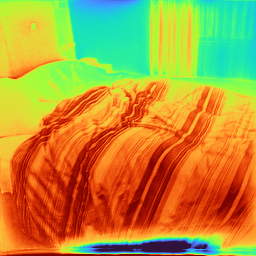}
    &
    \includegraphics[width=0.15\textwidth]{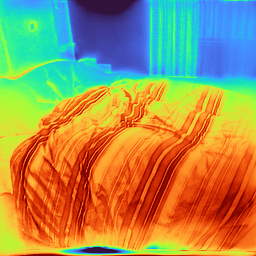}
    &
    \includegraphics[width=0.15\textwidth]{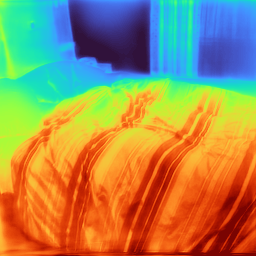}
    &
    \includegraphics[width=0.15\textwidth]{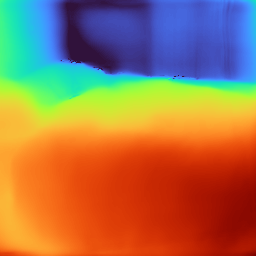}
    &
    \includegraphics[width=0.15\textwidth]{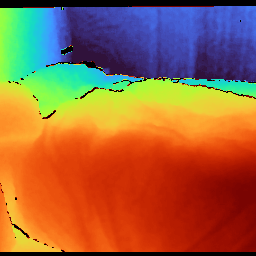}
    \\    
\end{tabular*}
\caption{\textbf{Qualitative comparison of rendered novel-view depth on RE10K (first row), ACID (second row), and ScanNet (last row).}}
\label{fig:qualitative_nvs_depth}
\end{figure*}

\noindent\textbf{Datasets and implementation details.}
Following~\cite{pixelsplat,mvsplat,noposplat}, we train on RealEstate10K~\cite{re10k} (RE10K) using the train/test split of~\cite{noposplat}. RE10K contains 67,477 training scenes and 7,289 testing scenes, with COLMAP~\cite{colmap} camera poses. We evaluate in-domain performance on RE10K and zero-shot generalization on ACID~\cite{acid} and ScanNet~\cite{scannet}. ACID contains outdoor aerial videos with COLMAP poses, while ScanNet provides RGB-D indoor sequences with camera motion and scene statistics that differ substantially from RE10K. Unless otherwise noted, all models are trained on RE10K only, without explicit depth supervision. Architecture and optimization details are provided in the supplementary material.

\subsection{Geometry Evaluation}
\label{subsec:geometry_eval}

Geometric veracity of the estimated 3D/2D Gaussian splats is the primary focus of this work. Traditionally, the geometry predicted by neural networks is evaluated by measuring the depth errors for the input views. However, input depths do not capture the interpolation capability of predicted Gaussians and are insensitive to the opacity, orientation, and scale. 
We propose a more holistic evaluation of the predicted scene structure by rendering multiple virtual depth maps from the reconstructed Gaussians and reporting Absolute Relative Error and thresholded depth accuracy. 
As we do not aim to extrapolate beyond the given view frustum, we use the same view-synthesis test set for depth evaluation. Virtual depth maps are rendered using the ground-truth relative pose w.r.t the first input frame, assuming perfectly aligned multi-view Gaussians. This puts pose-free methods at a severe disadvantage -- small pose-alignment errors amplify depth errors -- yet our pose-free method outperforms pose-aware counterparts by a large margin, as shown in \Cref{tab:scannet_depth_novel_source}.

\begin{figure*}[hbt!]
  \centering
\includegraphics[width=\textwidth]{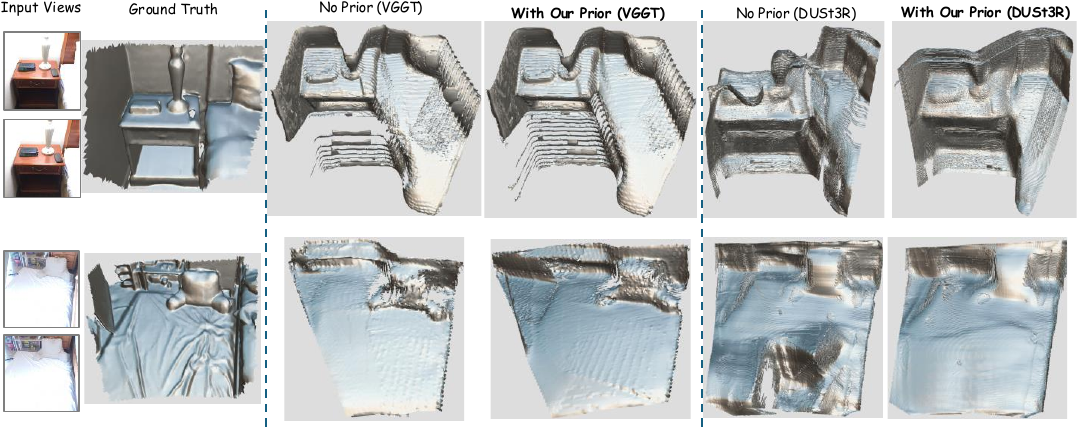}
\caption{
\textbf{Qualitative ablation of reconstructed meshes on ScanNet~\cite{scannet} (2 input views) using VGGT~\cite{VGGT} and DUSt3R~\cite{dust3r} backbones.} Our proposed priors consistently yield sharper, more complete, and less noisy geometry across both backbones.
}
\label{fig:mesh_reconstruction_scannet}
\end{figure*}

\newcommand{\groupcell}[1]{%
  \scriptsize
  \begin{tabular}[c]{@{}l@{}}#1\end{tabular}%
}
\begin{table}[t]
\centering
\caption{
\textbf{Depth estimation on ScanNet~\cite{scannet}: novel and source views.}
We report depth accuracy for rendered novel views and source views. 
DUSt3R~\cite{dust3r} is included as a supervised geometry reference and is therefore evaluated only on source-view depth.
Best results among generalizable splatting methods are shown in bold.
}
\label{tab:scannet_depth_novel_source}
\begin{adjustbox}{max width=\columnwidth}
\renewcommand{\arraystretch}{1.12}
\setlength{\tabcolsep}{0pt}
\small
\begin{tabular*}{\columnwidth}{@{\extracolsep{\fill}}llcccccc@{}}
\toprule
& \multirow{2}{*}{Method}
& \multicolumn{2}{c}{Abs Rel $\downarrow$}
& \multicolumn{2}{c}{$\delta_1{<}1.10 \uparrow$}
& \multicolumn{2}{c}{$\delta_1{<}1.25 \uparrow$} \\
\cmidrule(lr){3-4}\cmidrule(lr){5-6}\cmidrule(l){7-8}
&
& Novel & \src{Source}
& Novel & \src{Source}
& Novel & \src{Source} \\
\midrule
\midrule
\groupcell{\emph{Supervised}\\\emph{geometry}}
& DUSt3R~\cite{dust3r}
& \NA & \src{0.059}
& \NA & \src{0.886}
& \NA & \src{0.967} \\
\midrule
\midrule
\addlinespace[1pt]
\multirow{4}{*}{\groupcell{\emph{Pose-}\\\emph{required}}}
& pixelSplat~\cite{pixelsplat}
& 0.299 & \src{0.288}
& 0.552 & \src{0.553}
& 0.818 & \src{0.820} \\
& MVSplat~\cite{mvsplat}
& 0.189 & \src{0.132}
& 0.412 & \src{0.641}
& 0.745 & \src{0.891} \\
& FreeSplat~\cite{freesplat}
& 0.126 & \src{0.124}
& 0.556 & \src{0.556}
& 0.831 & \src{0.833} \\
& DepthSplat\cite{depthsplat}
& 0.135 & \src{0.105}
& 0.578 & \src{0.722}
& 0.864 & \src{0.914} \\
\midrule
\midrule
\addlinespace[1pt]
\multirow{5}{*}{\groupcell{\emph{Pose-}\\\emph{free}}}
& Splatt3R~\cite{splatt3r}
& 0.148 & \src{\NA}
& 0.546 & \src{\NA}
& 0.806 & \src{\NA} \\
& SelfSplat~\cite{selfsplat}
& 0.160 & \src{0.155}
& 0.502 & \src{0.460}
& 0.810 & \src{0.801} \\
& NoPoSplat~\cite{noposplat}
& 0.131 & \src{0.121}
& 0.554 & \src{0.662}
& 0.851 & \src{0.869} \\
& FLARE~\cite{flare} 
& 0.133 & \src{0.125} 
& 0.558 & \src{0.660}
& 0.853 & \src{0.870}\\
& \textbf{Ours}
& \textbf{0.090} & \src{\textbf{0.082}}
& \textbf{0.713} & \src{\textbf{0.740}}
& \textbf{0.916} & \src{\textbf{0.928}} \\
\bottomrule
\end{tabular*}
\end{adjustbox}
\end{table}

\Cref{tab:scannet_depth_novel_source} shows that G\textsuperscript{3}Splat substantially improves rendered depth on ScanNet. The gains are most visible for novel-view depth, where the representation must interpolate consistent geometry rather than merely reproduce source-view structure. In \Cref{fig:qualitative_nvs_depth}, MVSplat, DepthSplat, and NoPoSplat show texture-sensitive depth artifacts, while pixelSplat is noticeably noisy in textureless regions. Our method produces smoother and more plausible depth without requiring relative poses at inference.

For source-view depth, each baseline is evaluated using its strongest available depth estimate, either rendered from Gaussians or predicted directly by its depth-estimation head. For example, pixelSplat obtains its best source-view accuracy from rendered depth, whereas MVSplat and DepthSplat perform best using their network-predicted depths. Under this protocol, our method achieves the lowest AbsRel among the self-supervised splatting baselines and remains competitive in thresholded accuracy, while also producing substantially stronger novel-view geometry.

\begin{table}[t!]
\centering
\caption{
\textbf{Depth-estimation ablation on ScanNet~\cite{scannet} novel views.}
We ablate the proposed alignment and orientation priors for both 3DGS and 2DGS representations.
The ``Ref.'' column indicates optional test-time pose refinement following~\cite{noposplat}.
}
\label{tab:ablation_novel_depth}
\renewcommand{\arraystretch}{1.12}
\setlength{\tabcolsep}{4.1pt}
\small
\begin{tabular}{@{}lccc ccc@{}}
\toprule
\multirow{2}{*}{Rep.}
& \multicolumn{3}{c}{Ablated component}
& \multicolumn{3}{c}{Novel-view depth} \\
\cmidrule(lr){2-4}\cmidrule(l){5-7}
& $\mathcal{L}_{\mathrm{align}}$
& $\mathcal{L}_{\mathrm{orient}}$
& Ref.
& Abs Rel $\downarrow$
& $\delta_1{<}1.10 \uparrow$
& $\delta_1{<}1.25 \uparrow$ \\
\midrule
\multirow{8}{*}{3DGS}
& \xmark & \xmark & \xmark & 0.106 & 0.688 & 0.897 \\
& \xmark & \xmark & \cmark & 0.102 & 0.715 & 0.901 \\
\cmidrule(lr){2-7}
& \cmark & \xmark & \xmark & 0.097 & 0.701 & 0.907 \\
& \cmark & \xmark & \cmark & 0.089 & 0.729 & 0.920 \\
\cmidrule(lr){2-7}
& \xmark & \cmark & \xmark & 0.093 & 0.707 & 0.913 \\
& \xmark & \cmark & \cmark & 0.085 & 0.733 & 0.925 \\
\cmidrule(lr){2-7}
& \cmark & \cmark & \xmark & \textbf{0.090} & \textbf{0.713} & \textbf{0.916} \\
& \cmark & \cmark & \cmark & \textbf{0.083} & \textbf{0.738} & \textbf{0.928} \\
\midrule
\midrule
\multirow{8}{*}{2DGS}
& \xmark & \xmark & \xmark & 0.121 & 0.668 & 0.879 \\
& \xmark & \xmark & \cmark & 0.114 & 0.692 & 0.884 \\
\cmidrule(lr){2-7}
& \cmark & \xmark & \xmark & 0.107 & 0.684 & 0.894 \\
& \cmark & \xmark & \cmark & 0.099 & 0.713 & 0.908 \\
\cmidrule(lr){2-7}
& \xmark & \cmark & \xmark & 0.097 & 0.704 & 0.909 \\
& \xmark & \cmark & \cmark & 0.090 & 0.726 & 0.920 \\
\cmidrule(lr){2-7}
& \cmark & \cmark & \xmark & \textbf{0.094} & \textbf{0.715} & \textbf{0.916} \\
& \cmark & \cmark & \cmark & \textbf{0.082} & \textbf{0.743} & \textbf{0.931} \\
\bottomrule
\end{tabular}
\end{table}

The ablation in \Cref{tab:ablation_novel_depth} isolates the effect of the two proposed priors. The alignment loss $\mathcal{L}_{\text{align}}$ improves the consistency of Gaussian means, while the orientation loss $\mathcal{L}_{\text{orient}}$ directly stabilizes the Gaussian normal direction. The full objective gives the best or near-best results for both 3DGS and 2DGS variants. Importantly, $\mathcal{L}_{\text{align}}$ alone is not sufficient to learn stable Gaussian orientations; adding $\mathcal{L}_{\text{orient}}$ yields sharper rendered depth, color, and surface normals.

\paragraph{Mesh reconstruction.}
To evaluate whether the predicted splats support coherent surface recovery, we reconstruct meshes using only virtual novel views. For each ScanNet scene, we predict Gaussians from only source images, synthesize an interpolated camera trajectory between the sources, render per-view depth maps, and fuse them using TSDF-Fusion~\cite{tsdf-fusion}. We compare to ground-truth ScanNet meshes using accuracy, completeness, and Chamfer distance~\cite{guo2022neural,niceslam}.

\begin{figure*}[t!]
  \centering
\includegraphics[width=\textwidth]{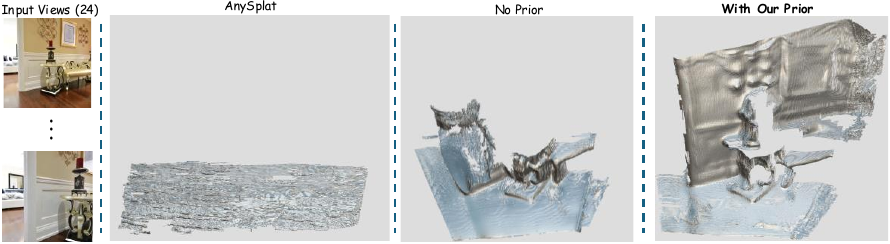}
\caption{
\textbf{Qualitative ablation of reconstructed meshes on RE10K~\cite{re10k} (24 input views; VGGT~\cite{VGGT} backbone).}
Our proposed priors yield more complete reconstructions and improve multi-view geometric consistency.
}
\label{fig:mesh_reconstruction_re10k}
\end{figure*}

\begin{table}[t]
\centering
\caption{
\textbf{Mesh reconstruction ablation on ScanNet~\cite{scannet}.}
Meshes are reconstructed by rendering virtual depth maps from the predicted Gaussians and fusing them with TSDF-Fusion~\cite{tsdf-fusion}.
The proposed geometric priors improve both DUSt3R- and VGGT-based splat predictors.
}
\label{tab:mesh_reconstruction_scannet}
\renewcommand{\arraystretch}{1.12}
\setlength{\tabcolsep}{0pt}
\small
\begin{tabular*}{0.9\columnwidth}{@{\extracolsep{\fill}}llccc@{}}
\toprule
Backbone & Our priors & Acc. $\downarrow$ & Comp. $\downarrow$ & Chamfer $\downarrow$ \\
\midrule
\multirow{2}{*}{DUSt3R~\cite{dust3r}}
& \xmark & 0.266 & 0.514 & 0.390 \\
& \cmark & \textbf{0.255} & \textbf{0.498} & \textbf{0.377} \\
\midrule
\multirow{2}{*}{VGGT~\cite{VGGT}}
& \xmark & 0.150 & 0.362 & 0.256 \\
& \cmark & \textbf{0.139} & \textbf{0.349} & \textbf{0.244} \\
\bottomrule
\end{tabular*}
\end{table}

\Cref{tab:mesh_reconstruction_scannet} shows that the priors improve both DUSt3R-style and VGGT-style variants. The qualitative results in \Cref{fig:mesh_reconstruction_scannet} show the same trend: with the proposed priors, reconstructed meshes are sharper, less noisy, and more complete across both backbones. This is a direct consequence of predicting Gaussians whose centers and local covariance structure are geometrically coherent rather than merely optimized to explain a few rendered images.

\begin{table*}[t]
\centering
\caption{\textbf{VGGT-based geometry comparison with AnySplat on ScanNet~\cite{scannet}.}
We compare rendered-depth accuracy and TSDF-Fusion mesh quality using VGGT-based splat predictors under sparse-view and multi-view evaluation settings.
``Views'' denotes the number of source images provided at inference to construct the Gaussian representation: the 2-view setting evaluates sparse reconstruction, while the 24-view setting evaluates multi-view consistency and aggregation.
Here, N and S denote novel-view and source-view rendered-depth evaluation, respectively, and CD denotes Chamfer distance.
AnySplat$^\dagger$~\cite{anysplat} fine-tunes a VGGT-based model using pseudo-depth supervision and additional training data, whereas our VGGT adaptation is trained with view synthesis and the proposed geometric priors.
Best results within each source-view setting are shown in bold. The multi-view setting narrows the rendered-depth gap, but the proposed priors continue to yield clear gains in TSDF-fused mesh quality, indicating improved global geometric consistency.
}
\label{tab:anysplat_vggt_geometry}
\renewcommand{\arraystretch}{1.12}
\setlength{\tabcolsep}{0pt}
\small
\begin{tabular*}{0.9\textwidth}{@{\extracolsep{\fill}}c l cccc ccc@{}}
\toprule
\multirow{2}{*}{Views}
& \multirow{2}{*}{Method}
& \multicolumn{4}{c}{Rendered depth}
& \multicolumn{3}{c}{Mesh reconstruction} \\
\cmidrule(lr){3-6}\cmidrule(l){7-9}
&
& Abs Rel (N) $\downarrow$
& Abs Rel (S) $\downarrow$
& $\delta_1$ (N) $\uparrow$
& $\delta_1$ (S) $\uparrow$
& Acc. $\downarrow$
& Comp. $\downarrow$
& CD $\downarrow$ \\
\midrule
\multirow{2}{*}{2}
& AnySplat$^\dagger$
& 0.054 & 0.053 & 0.871 & 0.872 
& 0.255 & 0.511 & 0.383 \\
& \textbf{Ours}
& \textbf{0.045} & \textbf{0.045} & \textbf{0.908} & \textbf{0.908}
& \textbf{0.139} & \textbf{0.349} & \textbf{0.244} \\
\midrule
\midrule
\multirow{2}{*}{24}
& AnySplat$^\dagger$
& 0.045 & 0.045 & 0.894 & 0.894 
& 0.130 & 0.365 & 0.248 \\
& \textbf{Ours}
& \textbf{0.041} & \textbf{0.041} & \textbf{0.913} & \textbf{0.912} 
& \textbf{0.087} & \textbf{0.311} & \textbf{0.199} \\
\bottomrule
\end{tabular*}
\end{table*}

\paragraph{VGGT-based adaptation and multi-view reconstruction.}
For the VGGT-based variant, we keep the original pose and point-cloud branches of VGGT~\cite{VGGT} and append a Gaussian decoder that predicts orientations, scales, opacities, and colors. The depth branch is not used for Gaussian supervision. Following VGGT-style splatting adaptations~\cite{anysplat}, we account for the intrinsic/extrinsic convention of the backbone and include a pseudo-pose consistency term to stabilize the camera branch. The key question is whether the proposed orientation and alignment priors remain useful when the underlying structure predictor is already strong.

\begin{figure}[b!]
\centering
\includegraphics[width=\columnwidth]{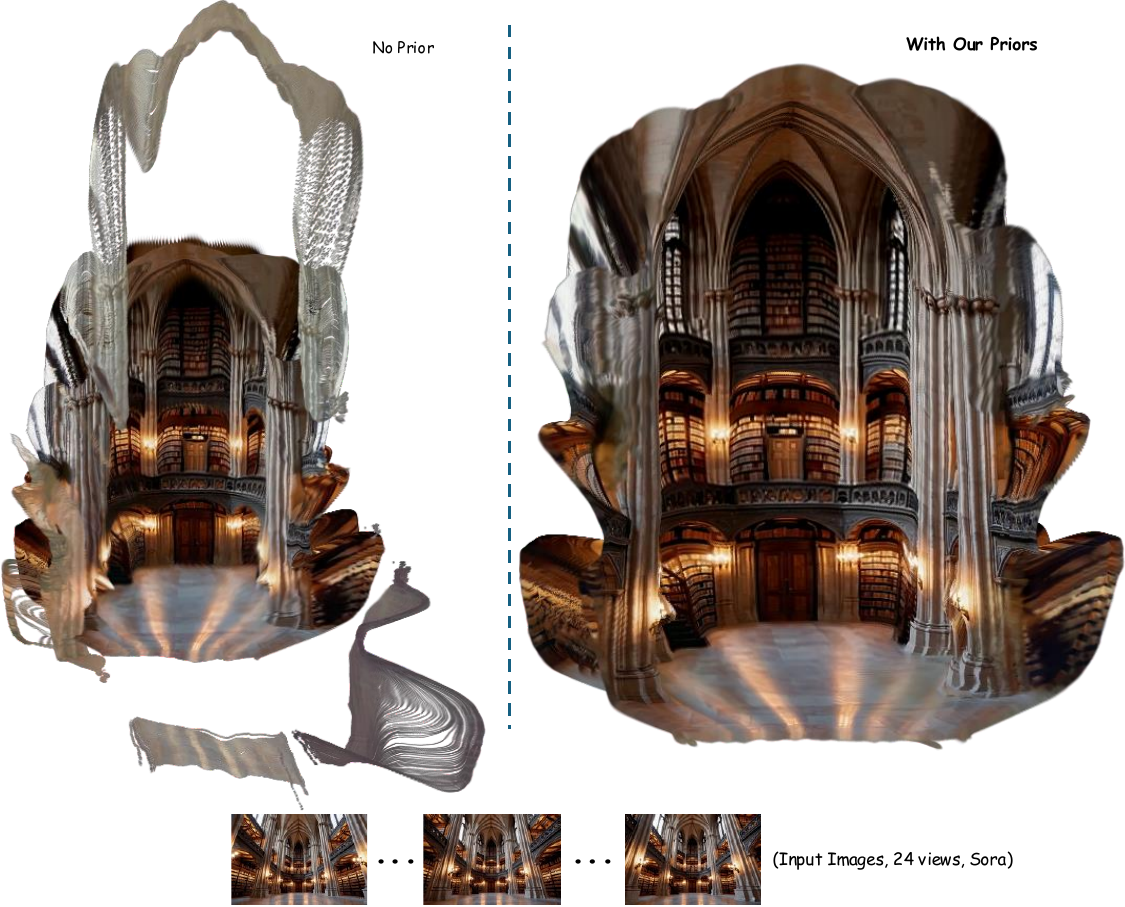}
\caption{\textbf{Qualitative ablation of reconstructed Gaussians on a Sora-generated video (VGGT backbone; model trained with two views and evaluated with 24 input views).} Prompt used to generate the video: ``A single unbroken orbital camera move through a vast, empty gothic library, with static architecture, medium-wide framing, warm steady lighting, and crisp sharp geometric details''.}
\label{fig:3dgs_reconstruction_sora}
\end{figure}

\begin{figure*}[t!]
  \centering
  \setlength{\tabcolsep}{2pt}

  \begin{minipage}[t]{0.58\textwidth}
    \centering
    \small \textbf{(a) RE10K, 24 input views}
    \vspace{2pt}

    \begin{tabular}{@{}c@{\hspace{3pt}}c@{}}
      {\small No priors} & {\small \emph{\textbf{With our priors}}} 
      \\
      \includegraphics[width=0.48\linewidth]{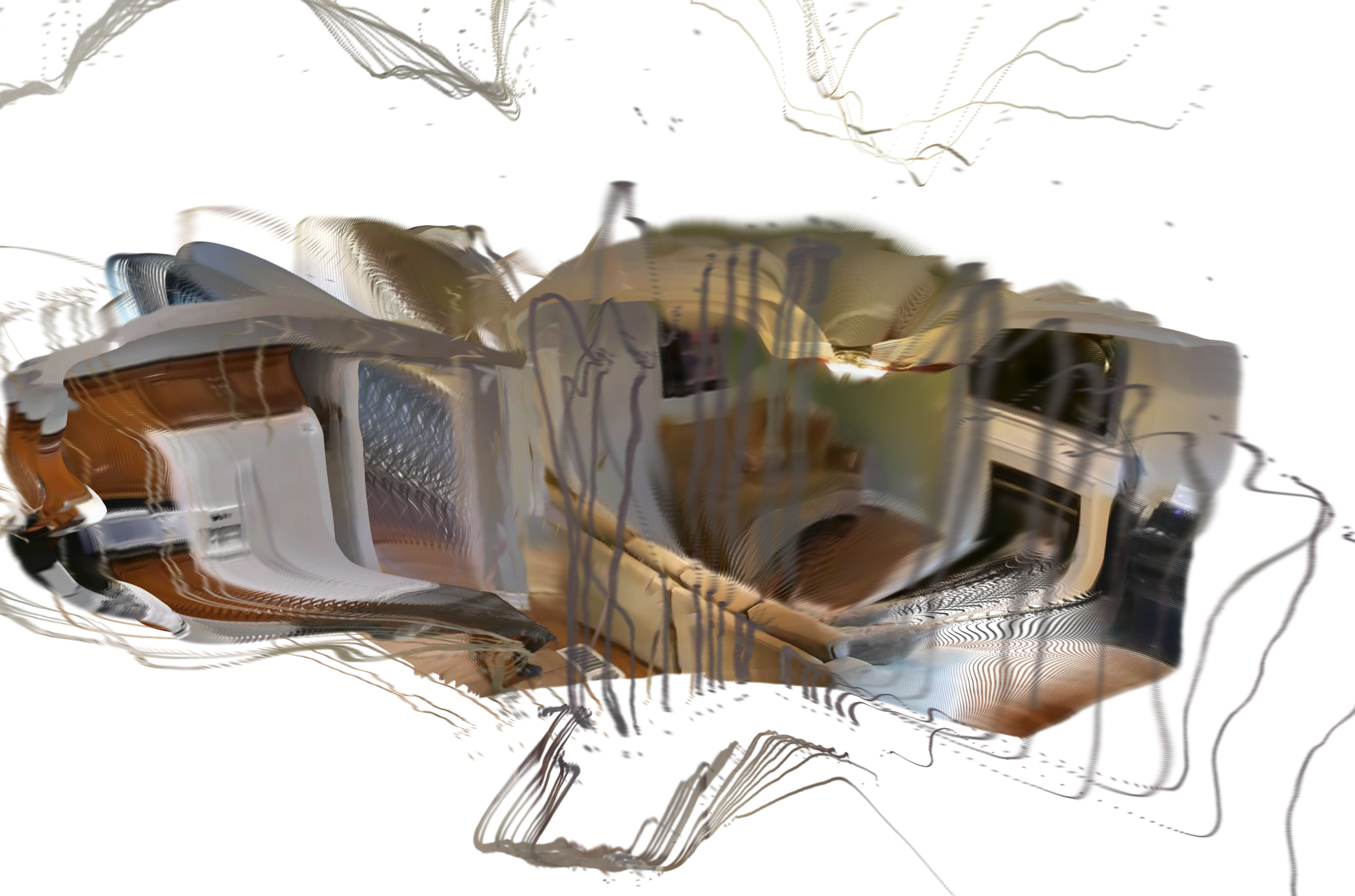} 
      &
      \includegraphics[width=0.55\linewidth,height=4cm]{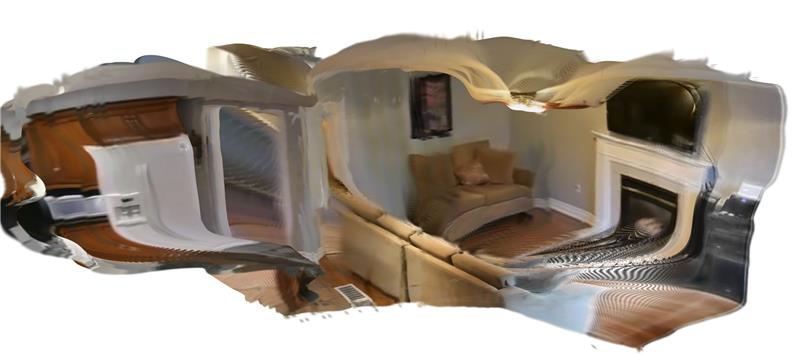}
    \end{tabular}
  \end{minipage}
  \hfill
  \begin{minipage}[t]{0.39\textwidth}
    \centering
    \small \textbf{(b) Tanks and Temples, 20 input views}
    \vspace{2pt}
    
    \begin{tabular}{@{}|c@{\hspace{3pt}}c@{}}
      {\small No priors} & {\small \emph{\textbf{With our priors}}} \\
      \includegraphics[width=0.48\linewidth,height=4cm]{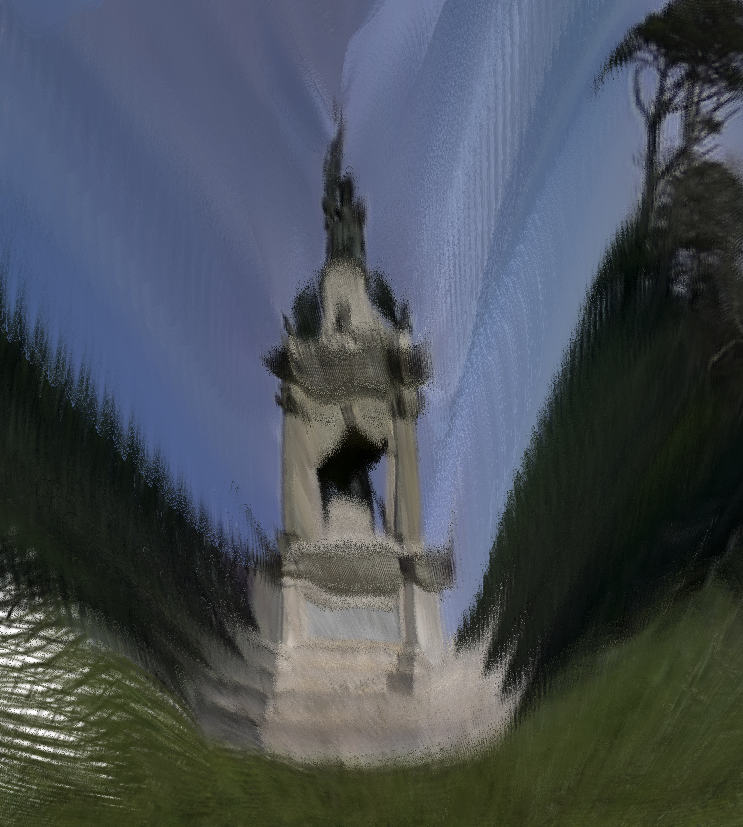}
      &
      \includegraphics[width=0.48\linewidth,height=4cm]{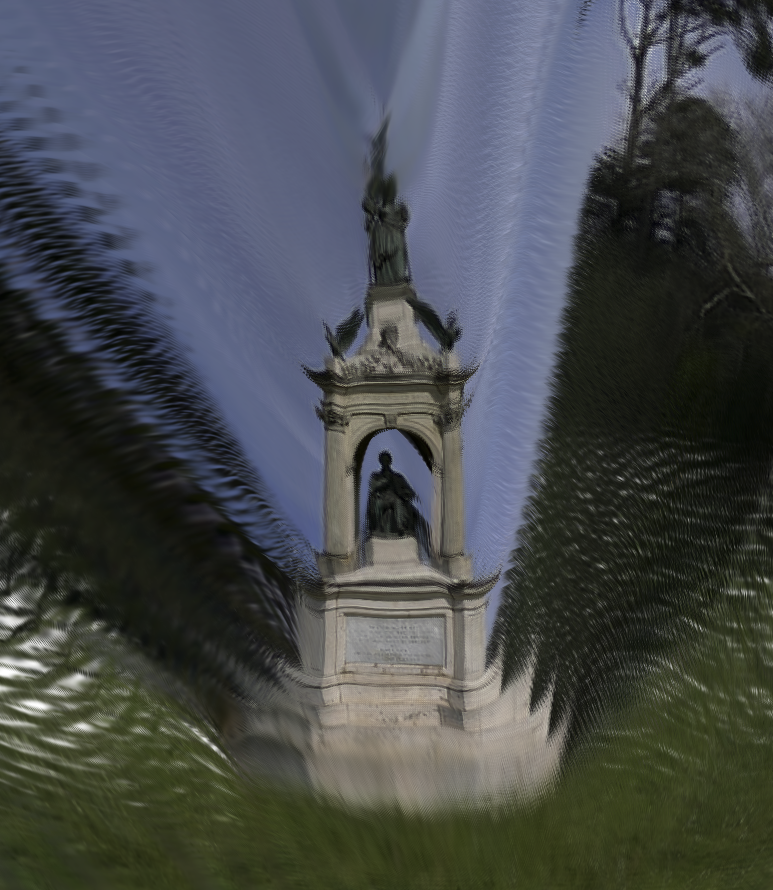}
    \end{tabular}
  \end{minipage}

  \caption{\textbf{Multi-view Gaussian reconstruction with a VGGT backbone.}
  We compare VGGT-based splat predictors trained with only view-synthesis supervision against the same models trained with the proposed geometric priors.
  On RE10K~\cite{re10k} and the out-of-domain Tanks and Temples~\cite{tnt} sequence, the prior-free model produces visibly inconsistent multi-view Gaussians: repeated structures do not align cleanly across frames, boundaries become smeared, and floating artifacts accumulate as more input views are fused.
  In contrast, the proposed priors encourage the predicted Gaussians from different views to agree in a common 3D frame, yielding a more coherent and geometrically consistent reconstruction without any post-processing or pose refinement.}
  \label{fig:3dgs_reconstruction_vggt_multiview}
\end{figure*}

Table~\ref{tab:anysplat_vggt_geometry} compares our VGGT-based adaptation with AnySplat~\cite{anysplat} under both sparse-view and multi-view evaluation settings. AnySplat fine-tunes a VGGT-based model using pseudo-depth supervision and additional training data, whereas our adaptation uses view-synthesis supervision together with the proposed geometric priors. We also note that AnySplat is trained on 24-views~\cite{anysplat}. In the two-view setting evaluation, our method improves rendered-depth accuracy and yields substantially better TSDF-fused meshes, reducing Chamfer distance from $0.383$ to $0.244$. With 24 source views, both methods benefit from additional observations, and the rendered-depth gap becomes smaller. Nevertheless, our priors continue to improve all reported metrics, with especially clear gains in mesh reconstruction: Chamfer distance decreases from $0.248$ to $0.199$, while both accuracy and completeness also improve. This suggests that the priors do not merely improve per-view depth quality; they also encourage the predicted Gaussians from different views to aggregate into a more coherent global surface. Notably, our two-view result is already competitive with the 24-view AnySplat setting on several metrics, indicating that the proposed losses provide useful geometric constraints even under sparse input. Overall, these results support the broader claim that the losses are not tied to a specific architecture, but provide reusable geometric constraints for splat predictors built on top of diverse multi-view structure backbones.

The qualitative results further highlight this multi-view consistency. Although the VGGT-based model is trained with two source views, it can be evaluated with many context views. \Cref{fig:mesh_reconstruction_re10k} shows a 24-view RE10K reconstruction: without the priors, misaligned per-frame Gaussians accumulate into floating or duplicated structures; with the priors, the splats consolidate into a more coherent surface. A similar effect is visible in \Cref{fig:3dgs_reconstruction_sora}, where the prior-free model leaves trails from cross-view misalignment on a Sora-generated sequence, while the prior-assisted model yields cleaner geometry and sharper textures. \Cref{fig:3dgs_reconstruction_vggt_multiview} shows the same trend on RE10K and demonstrates out-of-domain generalization on Tanks and Temples.

\begin{table*}[h]
\centering
\caption{\textbf{Relative pose estimation using PnP+RANSAC.}
We report AUC at multiple pose-error thresholds on RE10K~\cite{re10k} (in-domain) and on ScanNet~\cite{scannet} and ACID~\cite{acid} (cross-domain). The best overall result is shown in \textbf{bold}; the best result among pose-free splatting methods is \underline{underlined}. Methods marked with $\dagger$ are trained on additional data (e.g., ScanNet, ScanNet{\small ++}, ACID), and those marked with $\ddagger$ use extra supervision (e.g., ground-truth depth).}
\label{tab:pose_comparison_all}
\begin{adjustbox}{max width=\textwidth}
\setlength{\tabcolsep}{0pt}
\small
\begin{tabular*}{0.9\textwidth}{@{\extracolsep{\fill}}l l ccc ccc ccc@{}}
\toprule
&& \multicolumn{3}{c}{RE10K} & \multicolumn{3}{c}{ScanNet} & \multicolumn{3}{c}{ACID} \\
\cmidrule(lr){3-5} \cmidrule(lr){6-8} \cmidrule(lr){9-11}
& Method & 5$^\circ\uparrow$ & 10$^\circ\uparrow$ & 20$^\circ\uparrow$ & 5$^\circ\uparrow$ & 10$^\circ\uparrow$ & 20$^\circ\uparrow$ & 5$^\circ\uparrow$ & 10$^\circ\uparrow$ & 20$^\circ\uparrow$ \\
\midrule
\multirow{4}{*}{\shortstack[l]{\emph{Geometry/}\\\emph{pose baselines}}}
& CoPoNeRF$^\dag$~\cite{coponerf} & 0.161 & 0.362 & 0.575 & -- & -- & -- & 0.078 & 0.216 & 0.398 \\
& DUSt3R$^{\dag\ddag}$~\cite{dust3r} & 0.301 & 0.495 & 0.657 & 0.085 & 0.210 & 0.398 & 0.166 & 0.304 & 0.437 \\
& MASt3R$^{\dag\ddag}$~\cite{mast3r} & 0.372 & 0.561 & 0.709 & 0.083 & 0.200 & 0.381 & 0.234 & 0.396 & 0.541 \\
& RoMa$^{\dag\ddag}$~\cite{roma1} & 0.546 & 0.698 & 0.797 & \best{0.168} & \best{0.361} & \best{0.575} & \best{0.463} & \best{0.588} & \best{0.689} \\
\midrule
\midrule
\multirow{5}{*}{\shortstack[l]{\emph{Pose-free}\\\emph{splatting}}}
& Splatt3R$^\dag$~\cite{splatt3r} & 0.158 & 0.325 & 0.504 & 0.011 & 0.042 & 0.119 & 0.044 & 0.121 & 0.260 \\
& SelfSplat~\cite{selfsplat} & 0.030 & 0.083 & 0.180 & 0.030 & 0.098 & 0.254 & 0.064 & 0.153 & 0.283 \\
& FLARE~\cite{flare} & 0.025 & 0.122 & 0.324 & 0.116 & 0.262 & 0.467 & 0.091 & 0.236 & 0.441 \\ 
& NoPoSplat~\cite{noposplat} & 0.572 & 0.728 & 0.833 & 0.078 & 0.198 & 0.394 & 0.337 & 0.497 & 0.646 \\
& \best{Ours} & \best{0.629} & \best{0.770} & \best{0.858} & \categorybest{0.124} & \categorybest{0.282} & \categorybest{0.493} & \categorybest{0.404} & \categorybest{0.560} & \categorybest{0.689} \\
\bottomrule
\end{tabular*}
\end{adjustbox}
\end{table*}

\subsection{Relative Pose Evaluation}
\label{subsec:pose_eval}

Relative pose estimation provides an additional probe of the predicted scene structure. The predicted Gaussian means define dense 2D--3D correspondences between input pixels and the canonical scene frame. If these means are coherent and pixel-aligned, a classical PnP solver should recover accurate relative pose without relying on photometric pose refinement. Conversely, if the splats render plausible images but the means are geometrically inconsistent, PnP becomes unstable. We therefore use PnP-based relative pose as a proxy for the structural consistency of the predicted Gaussian means.

\begin{table}[t]
\centering
\caption{
\textbf{Pose-error ablation using least-squares PnP.}
We report translation and rotation errors on RE10K~\cite{re10k} and ScanNet~\cite{scannet}.
Translation error is the scale-invariant angular error between normalized translation vectors, and rotation error is measured in degrees.
Unlike PnP+RANSAC, least-squares PnP does not reject outliers, making this evaluation a direct stress test of the predicted Gaussian structure.
Best results within each representation are shown in bold.
}
\label{tab:pose_comparison_details}
\renewcommand{\arraystretch}{1.12}
\setlength{\tabcolsep}{0pt}
\small
\begin{tabular*}{0.9\columnwidth}{@{\extracolsep{\fill}}l c cc cc@{}}
\toprule
\multirow{2}{*}{Rep.}
& \multirow{2}{*}{Our priors}
& \multicolumn{2}{c}{RE10K}
& \multicolumn{2}{c}{ScanNet} \\
\cmidrule(lr){3-4}\cmidrule(l){5-6}
&
& Trans. $\downarrow$
& Rot. $\downarrow$
& Trans. $\downarrow$
& Rot. $\downarrow$ \\
\midrule
\multirow{2}{*}{3DGS}
& \xmark
& 27.423 & 37.560
& 56.839 & 144.950 \\
& \cmark
& \textbf{10.594} & \textbf{18.587}
& \textbf{30.848} & \textbf{55.294} \\
\midrule
\multirow{2}{*}{2DGS}
& \xmark
& 29.693 & 47.225
& 54.057 & 82.707 \\
& \cmark
& \textbf{8.495} & \textbf{17.420}
& \textbf{31.474} & \textbf{60.480} \\
\bottomrule
\end{tabular*}
\end{table}
%

\begin{table*}[t!]
\centering
\caption{
\textbf{Novel-view synthesis ablation on RE10K~\cite{re10k}, ACID~\cite{acid}, and ScanNet~\cite{scannet}.}
RE10K results are averaged over the small-, medium-, and large-overlap regimes of~\cite{noposplat}; ACID and ScanNet are zero-shot evaluations.
All results are obtained without target-pose optimization.
The proposed geometric priors are designed to improve structural consistency rather than directly optimize image quality; this ablation shows that they preserve, and in most cases slightly improve, novel-view synthesis performance across both in-domain and zero-shot settings.
Full comparisons with prior novel-view synthesis baselines are provided in the supplementary material.}
\label{tab:nvs_ablation_re10k_ood}
\renewcommand{\arraystretch}{1.12}
\small
\begin{tabular*}{0.9\textwidth}{@{\extracolsep{\fill}}lc ccc ccc ccc@{}}
\toprule
\multirow{2}{*}{Rep.}
& \multirow{2}{*}{Our priors}
& \multicolumn{3}{c}{RE10K}
& \multicolumn{3}{c}{ACID}
& \multicolumn{3}{c}{ScanNet} \\
\cmidrule(lr){3-5}\cmidrule(lr){6-8}\cmidrule(l){9-11}
&
& PSNR $\uparrow$ & SSIM $\uparrow$ & LPIPS $\downarrow$
& PSNR $\uparrow$ & SSIM $\uparrow$ & LPIPS $\downarrow$
& PSNR $\uparrow$ & SSIM $\uparrow$ & LPIPS $\downarrow$ \\
\midrule
\multirow{2}{*}{3DGS}
& \xmark
& 23.244 & 0.778 & 0.187
& 23.379 & 0.684 & 0.237
& 21.069 & 0.646 & 0.269 \\
& \cmark
& \textbf{23.417} & \textbf{0.783} & \textbf{0.185}
& \textbf{23.763} & \textbf{0.700} & \textbf{0.236}
& \textbf{21.137} & \textbf{0.648} & 0.269 \\
\midrule
\midrule
\multirow{2}{*}{2DGS}
& \xmark
& 23.407 & 0.782 & 0.188
& 23.825 & 0.700 & 0.235
& 21.167 & 0.648 & 0.267 \\
& \cmark
& \textbf{23.504} & \textbf{0.787} & \textbf{0.184}
& \textbf{23.827} & \textbf{0.701} & 0.235
& \textbf{21.168} & \textbf{0.650} & \textbf{0.266} \\
\bottomrule
\end{tabular*}
\end{table*}

\Cref{tab:pose_comparison_all} compares PnP+RANSAC pose estimation on RE10K, ScanNet, and ACID. CoPoNeRF~\cite{coponerf} is trained with explicit pose supervision. DUSt3R~\cite{dust3r} and MASt3R~\cite{mast3r} use large-scale supervised 3D regression data, and RoMa~\cite{roma1} is trained with joint depth-and-pose supervision on MegaDepth and ScanNet. In contrast, our model is trained on RE10K with view-synthesis supervision and the proposed geometric priors. Despite this weaker supervision, it outperforms pose-free splatting baselines across all datasets and remains competitive with geometry- and pose-supervised methods. The only consistent exception is RoMa on ScanNet, where RoMa benefits from explicit training on ScanNet.
Compared with NoPoSplat, the gains are especially clear in the PnP+RANSAC setting, where no photometric refinement is used. This supports the central claim that the priors improve the geometry of the predicted Gaussian means, rather than merely improving the renderability of a particular target view. Photometric test-time refinement is reported in the supplementary material; the main table focuses on PnP+RANSAC because it more directly reflects the geometric quality of the predicted splats.

\Cref{tab:pose_comparison_details} further removes RANSAC and evaluates least-squares PnP. This setting deliberately exposes outliers and therefore provides a stricter probe of the raw Gaussian means. The proposed priors significantly reduce both translation and rotation error for 3DGS and 2DGS representations. The translation error is scale-invariant and is computed as the angular difference between normalized predicted and ground-truth translation vectors.

\subsection{Novel-View Synthesis Evaluation}
\label{subsec:nvs_eval}

Novel-view synthesis is not the main focus of this work, but it remains an important sanity check: the proposed geometric priors should improve structure without degrading image quality. We therefore report image-synthesis metrics as an ablation of the proposed priors rather than emphasizing a baseline ranking. For pose-free methods, target views are rendered directly from all reconstructed Gaussian splats at a fixed target pose relative to the first input image. We do not optimize camera poses for target-view image synthesis, since pose refinement can mask geometric inconsistencies and effectively use the target view to compensate for an inaccurate reconstruction~\cite{barf,nerfmm,chng2024invert,garg2024direct}.

\Cref{tab:nvs_ablation_re10k_ood} shows that the geometric priors preserve, and slightly improve, novel-view synthesis quality across in-domain and zero-shot settings. The gains are modest relative to the improvements in depth, mesh, and pose estimation, which is expected: the priors are designed to resolve ambiguous Gaussian orientations, scales, and alignment rather than to directly optimize perceptual image metrics. This result is therefore best interpreted as evidence that the geometric improvements do not come at the expense of the original view-synthesis objective.

%% file: sec/5_conclusion.tex
\section{Conclusion}
\label{sec:conclusion}

We presented G\textsuperscript{3}Splat, a geometry-consistent framework for generalizable Gaussian splatting. The main observation is that view-synthesis supervision alone is under-constrained for predicting Gaussian orientations, scales, and cross-view alignment: splats can render plausible images while remaining geometrically degenerate. We addressed this ambiguity with two simple and architecture-agnostic priors, one aligning Gaussian orientations with local surface normals and the other keeping Gaussian centers pixel-aligned. Across DUSt3R-style and VGGT-style backbones, and across both 3DGS and surfel-like 2DGS variants, these priors improve rendered depth, mesh reconstruction, and relative pose estimation while preserving strong novel-view synthesis quality. The results suggest that future generalizable splatting methods should evaluate and optimize the geometric consistency of the predicted splats, rather than relying on image reconstruction alone.

%% file: sec/appendix.tex
\appendices
\section{Supplementary Overview}
\label{sec:supp_overview}

These appendices provide additional evaluations, supporting ablations, implementation details, and evaluation protocols for G\textsuperscript{3}Splat. The emphasis remains on geometric consistency across generalizable Gaussian splatting backbones, while avoiding repetition of the primary experimental discussion. 

Specifically, \Cref{sec:supp_additional_evals} reports additional DUSt3R-based geometry ablations, complete novel-view synthesis comparisons, full source/novel-view depth ablations, and additional qualitative comparisons. \Cref{sec:supp_architectures} describes the DUSt3R- and VGGT-based instantiations of the proposed method. \Cref{sec:supp_implementation_details} gives training and evaluation details, including the VGGT-specific conventions used when attaching the Gaussian decoder. \Cref{sec:supp_rendered_normal} analyzes rendered normal--depth consistency for 2DGS and explains why the direct orientation prior used by G\textsuperscript{3}Splat is preferred. \Cref{sec:supp_depth_renderer,sec:supp_pose_refinement,sec:supp_mesh_evaluation_protocol} describe depth rendering, optional test-time pose refinement, and mesh reconstruction/evaluation protocols, respectively.

\section{Additional Evaluations}
\label{sec:supp_additional_evals}

\subsection{Additional DUSt3R-Based Geometry Ablations}
\label{subsec:additional_eval_dust3r}

We provide further qualitative comparisons of novel-view depth maps rendered by our method on RE10K~\cite{re10k} and ScanNet~\cite{scannet} in \Cref{fig:supp_qualitative_nv_depth}. 

We also report more detailed ablations of our proposed priors for both pose and depth estimation. For pose, \Cref{tab:ablation_pose} evaluates our models on RE10K~\cite{re10k}, ScanNet~\cite{scannet}, and ACID~\cite{acid} under three schemes: PnP (least squares), PnP+RANSAC, and test-time photometric refinement (see \Cref{sec:supp_pose_refinement} for details). For depth, \Cref{tab:ablation_depth} reports novel- and source-view depth error and accuracy on ScanNet~\cite{scannet}.

\begin{figure*}[hbt!]
  \centering
\includegraphics[width=\textwidth]{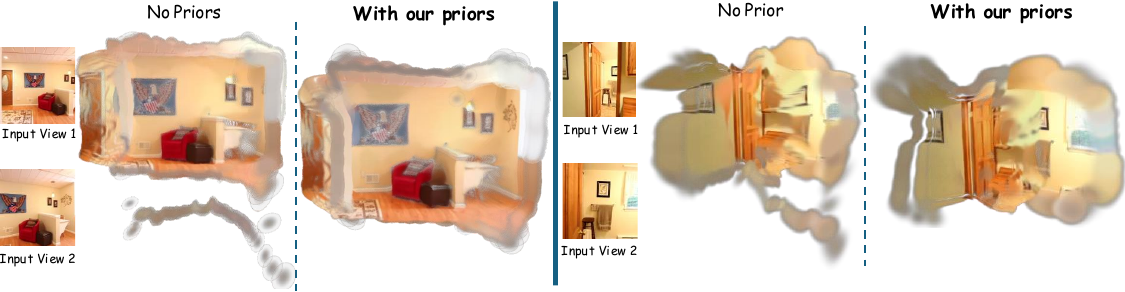}
\caption{\textbf{Qualitative ablation of reconstructed Gaussians on RE10K~\cite{re10k} (DUSt3R backbone, 2 input views).}}
\label{fig:3dgs_reconstruction_re10k}
\end{figure*}

Overall, the proposed priors yield substantial gains in geometric quality. In particular, under the PnP (least-squares) setting they almost double the pose AUC on RE10K (in-domain) and ACID (cross-domain), with even larger relative improvements on ScanNet, indicating fewer outliers and more reliable Gaussian means. Similarly, on ScanNet we observe more than a 20\% reduction in absolute relative depth error for both novel and source views when both priors are enabled. These geometric improvements are also apparent in the reconstructed Gaussians on RE10K (\Cref{fig:3dgs_reconstruction_re10k}), which exhibit cleaner geometry and markedly fewer floating artifacts.

\begin{figure*}[t]
  \centering
  \includegraphics[width=\textwidth]{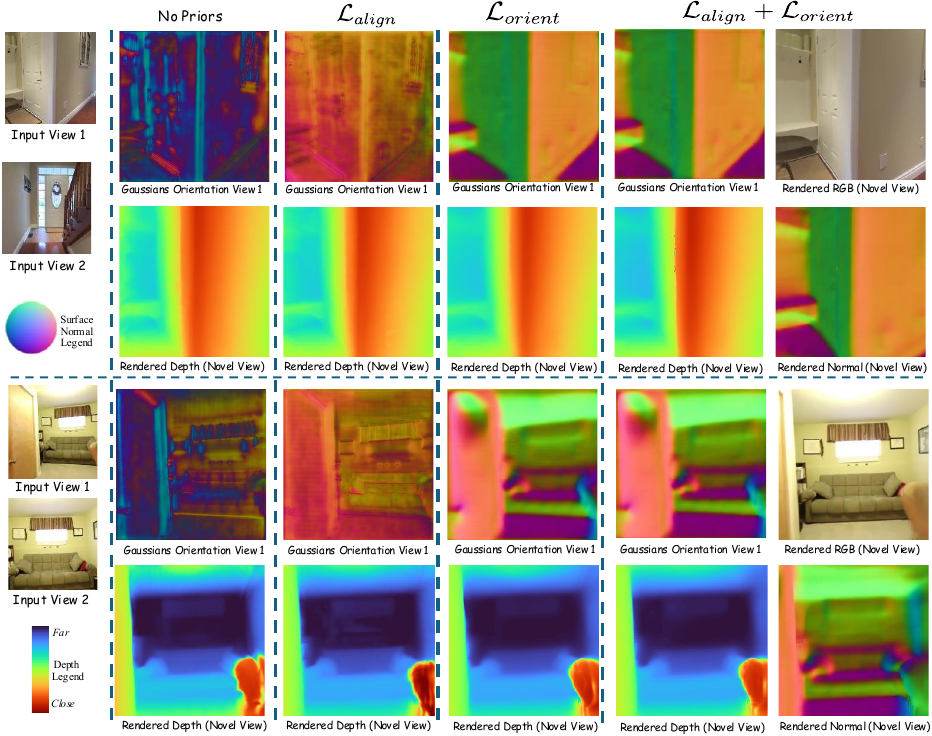}
  \caption{\textbf{Qualitative ablation of the losses.}
  We visualize the learned Gaussian orientations (from the first input view) along with rendered novel-view depth, color, and surface normals, on RE10K~\cite{re10k}.
  As shown, using $\mathcal{L}_{\text{align}}$ alone is insufficient to learn reliable Gaussian orientations; adding $\mathcal{L}_{\text{orient}}$ encourages geometry-consistent orientations and, together, they produce more accurate rendered depth.}
  \label{fig:loss_ablations}
\end{figure*}

The ablations show that $\mathcal{L}_{\text{align}}$ alone is insufficient to learn stable Gaussian orientations. Adding $\mathcal{L}_{\text{orient}}$ enforces geometry-consistent normals and yields sharper novel-view renderings, as illustrated in \Cref{fig:loss_ablations}.

\begin{figure*}[t!]
\centering
    \includegraphics[width=\textwidth]{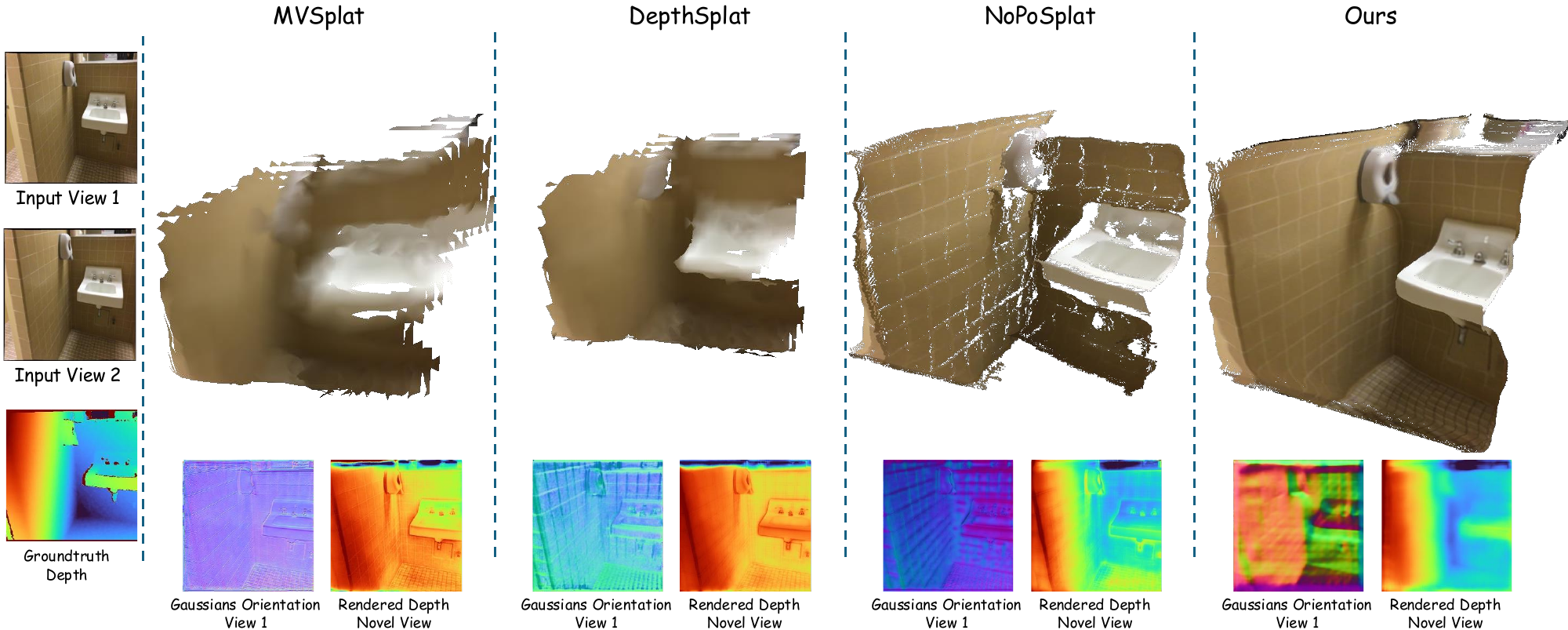}
    \includegraphics[width=\textwidth,trim={0 0 0 20},clip]{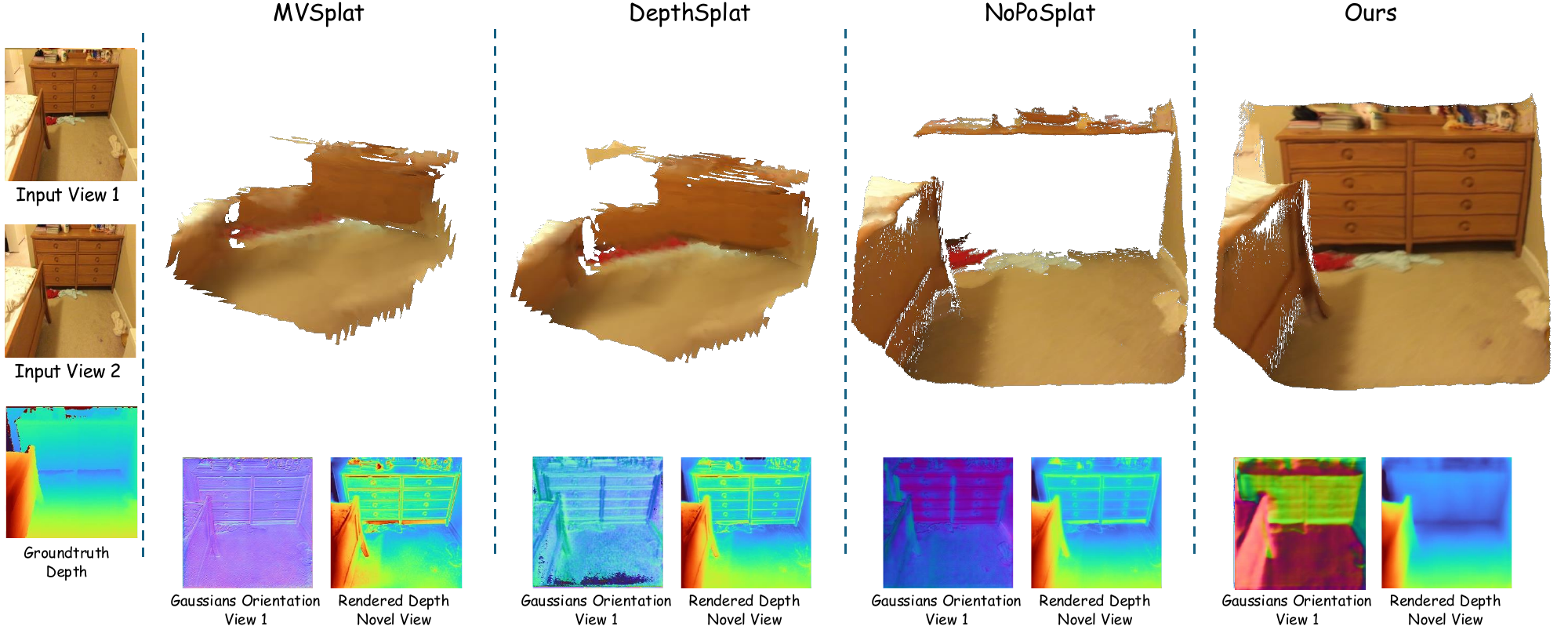}
    \includegraphics[width=\textwidth,trim={0 0 0 15},clip]{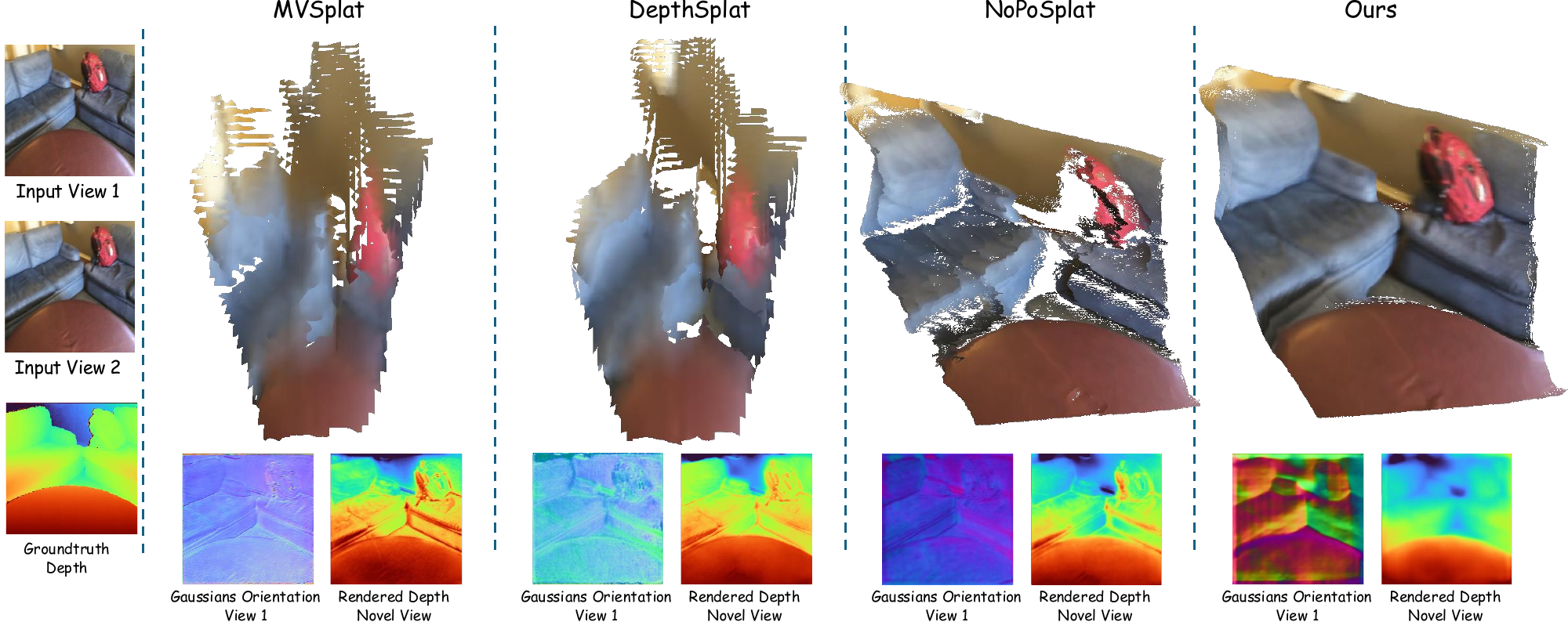}
\caption{\textbf{Qualitative comparison of mesh reconstruction on ScanNet~\cite{scannet} (DUSt3R backbone, 2 input views).}
For each scene, we show the two input context views, the textured mesh reconstructed by fusing \emph{virtual} rendered depth maps via TSDF-Fusion~\cite{tsdf-fusion}, the Gaussian normals for the first input view, and the ground-truth and rendered depth from a novel (virtual) viewpoint.
Baselines exhibit inaccurate rendered depth and normals; when fused, these inconsistencies lead to holes and deformed regions in the reconstructed meshes.}
\label{fig:supp_qualitative_mesh_scannet}
\end{figure*}

Finally, \Cref{fig:supp_qualitative_mesh_scannet} compares mesh reconstructions from two input views on ScanNet~\cite{scannet}. We evaluate pose-required baselines MVSplat~\cite{mvsplat} and DepthSplat~\cite{depthsplat}, the pose-free NoPoSplat~\cite{noposplat}, and our method. Meshes are reconstructed by fusing \emph{virtual} (novel-view) rendered depth maps via TSDF-Fusion~\cite{tsdf-fusion} (see \Cref{sec:supp_mesh_evaluation_protocol} for details). For each method, we visualize Gaussian orientations (surface normals) for the first input view alongside the rendered depth for a novel view and its ground-truth depth. Our approach consistently produces geometrically coherent meshes, whereas competing methods' inconsistent novel-view depths often lead to deformed planar regions (rows one and three) and large holes (row two), losing fine scene detail.

\begin{table*}[t]
\centering
\caption{
\textbf{Pose-estimation ablation on RE10K, ScanNet, and ACID.}
We report AUC at $5^\circ$, $10^\circ$, and $20^\circ$ on RE10K~\cite{re10k} for in-domain evaluation, and on ScanNet~\cite{scannet} and ACID~\cite{acid} for cross-domain evaluation.
We compare 3DGS and 2DGS variants under three pose-estimation schemes: least-squares PnP, PnP+RANSAC, and optional photometric test-time refinement following the same synthesis loss used during training.
Best results within each pose-estimation scheme are shown in bold.}
\label{tab:ablation_pose}
\renewcommand{\arraystretch}{1.10}
\setlength{\tabcolsep}{0pt}
\small
\begin{tabular*}{0.9\textwidth}{@{\extracolsep{\fill}}llcc ccc ccc ccc@{}}
\toprule
\multirow{2}{*}{\schemecell{Pose \\scheme}}
& \multirow{2}{*}{Rep.}
& \multicolumn{2}{c}{Priors}
& \multicolumn{3}{c}{RE10K}
& \multicolumn{3}{c}{ScanNet}
& \multicolumn{3}{c}{ACID} \\
\cmidrule(lr){3-4}
\cmidrule(lr){5-7}
\cmidrule(lr){8-10}
\cmidrule(l){11-13}
&
& $\mathcal{L}_{\mathrm{align}}$
& $\mathcal{L}_{\mathrm{orient}}$
& $5^\circ$ & $10^\circ$ & $20^\circ$
& $5^\circ$ & $10^\circ$ & $20^\circ$
& $5^\circ$ & $10^\circ$ & $20^\circ$ \\
\midrule

\multirow{8}{*}{\schemecell{\emph{PnP}\\\emph{(LS)}}}
& \multirow{4}{*}{3DGS}
& \xmark & \xmark
& 0.296 & 0.437 & 0.570
& 0.019 & 0.050 & 0.104
& 0.189 & 0.279 & 0.378 \\
&
& \cmark & \xmark
& 0.388 & 0.558 & 0.692
& 0.029 & 0.089 & 0.199
& 0.250 & 0.366 & 0.476 \\
&
& \xmark & \cmark
& 0.477 & 0.642 & 0.753
& \textbf{0.068} & \textbf{0.141} & \textbf{0.221}
& 0.296 & 0.414 & 0.508 \\
&
& \cmark & \cmark
& 0.482 & 0.645 & 0.753
& 0.052 & 0.121 & 0.209
& 0.301 & 0.416 & 0.511 \\
\cmidrule(lr){2-13}
& \multirow{4}{*}{2DGS}
& \xmark & \xmark
& 0.223 & 0.344 & 0.487
& 0.008 & 0.026 & 0.076
& 0.154 & 0.237 & 0.339 \\
&
& \cmark & \xmark
& 0.338 & 0.519 & 0.668
& 0.031 & 0.083 & 0.176
& 0.229 & 0.341 & 0.453 \\
&
& \xmark & \cmark
& 0.194 & 0.252 & 0.314
& 0.015 & 0.035 & 0.055
& 0.049 & 0.063 & 0.093 \\
&
& \cmark & \cmark
& \textbf{0.526} & \textbf{0.679} & \textbf{0.774}
& 0.051 & 0.126 & 0.210
& \textbf{0.327} & \textbf{0.442} & \textbf{0.529} \\
\midrule
\midrule

\multirow{8}{*}{\schemecell{\emph{PnP}\\\emph{+RANSAC}}}
& \multirow{4}{*}{3DGS}
& \xmark & \xmark
& 0.572 & 0.728 & 0.833
& 0.078 & 0.198 & 0.394
& 0.337 & 0.497 & 0.646 \\
&
& \cmark & \xmark
& 0.594 & 0.742 & 0.840
& 0.090 & 0.222 & 0.431
& 0.344 & 0.506 & 0.650 \\
&
& \xmark & \cmark
& 0.600 & 0.746 & 0.845
& \textbf{0.132} & \textbf{0.298} & 0.491
& 0.367 & 0.533 & 0.674 \\
&
& \cmark & \cmark
& \textbf{0.629} & \textbf{0.770} & \textbf{0.858}
& 0.124 & 0.282 & \textbf{0.493}
& \textbf{0.404} & \textbf{0.560} & \textbf{0.689} \\
\cmidrule(lr){2-13}
& \multirow{4}{*}{2DGS}
& \xmark & \xmark
& 0.588 & 0.737 & 0.832
& 0.085 & 0.223 & 0.432
& 0.344 & 0.513 & 0.659 \\
&
& \cmark & \xmark
& 0.619 & 0.759 & 0.849
& 0.120 & 0.279 & 0.471
& 0.382 & 0.540 & 0.674 \\
&
& \xmark & \cmark
& 0.592 & 0.743 & 0.836
& 0.099 & 0.241 & 0.448
& 0.374 & 0.535 & 0.672 \\
&
& \cmark & \cmark
& \textbf{0.629} & 0.768 & 0.856
& 0.128 & 0.281 & 0.477
& 0.387 & 0.546 & 0.682 \\
\midrule
\midrule

\multirow{8}{*}{\schemecell{\emph{Photometric}\\\emph{refinement}}}
& \multirow{4}{*}{3DGS}
& \xmark & \xmark
& 0.672 & 0.791 & 0.868
& 0.109 & 0.256 & 0.463
& 0.456 & 0.593 & 0.705 \\
&
& \cmark & \xmark
& 0.680 & 0.797 & 0.871
& 0.129 & 0.284 & 0.513
& 0.460 & 0.596 & 0.709 \\
&
& \xmark & \cmark
& 0.684 & 0.801 & 0.874
& 0.144 & 0.318 & 0.527
& 0.469 & 0.604 & 0.718 \\
&
& \cmark & \cmark
& 0.684 & 0.801 & \textbf{0.875}
& 0.148 & 0.326 & 0.540
& 0.466 & 0.598 & 0.713 \\
\cmidrule(lr){2-13}
& \multirow{4}{*}{2DGS}
& \xmark & \xmark
& 0.672 & 0.788 & 0.859
& 0.129 & 0.298 & 0.494
& 0.460 & 0.599 & 0.713 \\
&
& \cmark & \xmark
& 0.681 & 0.799 & 0.870
& 0.136 & 0.311 & 0.512
& 0.474 & 0.607 & 0.718 \\
&
& \xmark & \cmark
& 0.675 & 0.793 & 0.869
& 0.130 & 0.301 & 0.503
& 0.466 & 0.601 & 0.714 \\
&
& \cmark & \cmark
& \textbf{0.686} & \textbf{0.802} & \textbf{0.875}
& \textbf{0.153} & \textbf{0.334} & \textbf{0.541}
& \textbf{0.478} & \textbf{0.609} & \textbf{0.723} \\
\bottomrule
\end{tabular*}
\end{table*}

\begin{table*}[t]
\centering
\caption{
\textbf{Depth-estimation ablation on ScanNet~\cite{scannet}: novel and source views.}
We ablate the alignment and orientation priors for both 3DGS and 2DGS representations.
The ``Ref.'' column denotes optional test-time pose refinement following~\cite{noposplat}.}
\label{tab:ablation_depth}
\renewcommand{\arraystretch}{1.12}
\setlength{\tabcolsep}{0pt}
\small
\begin{tabular*}{0.9\textwidth}{@{\extracolsep{\fill}}lccc ccc ccc@{}}
\toprule
\multirow{2}{*}{Rep.}
& \multicolumn{3}{c}{Ablated component}
& \multicolumn{3}{c}{Novel view}
& \multicolumn{3}{c}{Source view} \\
\cmidrule(lr){2-4}\cmidrule(lr){5-7}\cmidrule(l){8-10}
& $\mathcal{L}_{\mathrm{align}}$
& $\mathcal{L}_{\mathrm{orient}}$
& Ref.
& Abs Rel $\downarrow$
& $\delta_1{<}1.10 \uparrow$
& $\delta_1{<}1.25 \uparrow$
& Abs Rel $\downarrow$
& $\delta_1{<}1.10 \uparrow$
& $\delta_1{<}1.25 \uparrow$ \\
\midrule
\multirow{8}{*}{3DGS}
& \xmark & \xmark & \xmark
& 0.106 & 0.688 & 0.897
& 0.105 & 0.689 & 0.897 \\
& \xmark & \xmark & \cmark
& 0.102 & 0.715 & 0.901
& 0.097 & 0.707 & 0.905 \\
\cmidrule(lr){2-10}
& \cmark & \xmark & \xmark
& 0.097 & 0.701 & 0.907
& 0.089 & 0.729 & 0.918 \\
& \cmark & \xmark & \cmark
& 0.089 & 0.729 & 0.920
& 0.086 & 0.740 & 0.923 \\
\cmidrule(lr){2-10}
& \xmark & \cmark & \xmark
& 0.093 & 0.707 & 0.913
& 0.085 & 0.733 & 0.925 \\
& \xmark & \cmark & \cmark
& 0.085 & 0.733 & 0.925
& 0.083 & 0.742 & 0.928 \\
\cmidrule(lr){2-10}
& \cmark & \cmark & \xmark
& \textbf{0.090} & \textbf{0.713} & \textbf{0.916}
& \textbf{0.082} & \textbf{0.740} & \textbf{0.928} \\
& \cmark & \cmark & \cmark
& \textbf{0.083} & \textbf{0.738} & \textbf{0.928}
& \textbf{0.080} & \textbf{0.747} & \textbf{0.930} \\
\midrule
\midrule
\multirow{8}{*}{2DGS}
& \xmark & \xmark & \xmark
& 0.121 & 0.668 & 0.879
& 0.118 & 0.665 & 0.875 \\
& \xmark & \xmark & \cmark
& 0.114 & 0.692 & 0.884
& 0.105 & 0.705 & 0.894 \\
\cmidrule(lr){2-10}
& \cmark & \xmark & \xmark
& 0.107 & 0.684 & 0.894
& 0.100 & 0.712 & 0.904 \\
& \cmark & \xmark & \cmark
& 0.099 & 0.713 & 0.908
& 0.096 & 0.723 & 0.911 \\
\cmidrule(lr){2-10}
& \xmark & \cmark & \xmark
& 0.097 & 0.704 & 0.909
& 0.090 & 0.727 & 0.920 \\
& \xmark & \cmark & \cmark
& 0.090 & 0.726 & 0.920
& 0.090 & 0.735 & 0.922 \\
\cmidrule(lr){2-10}
& \cmark & \cmark & \xmark
& \textbf{0.094} & \textbf{0.715} & \textbf{0.916}
& \textbf{0.086} & \textbf{0.736} & \textbf{0.925} \\
& \cmark & \cmark & \cmark
& \textbf{0.082} & \textbf{0.743} & \textbf{0.931}
& \textbf{0.079} & \textbf{0.752} & \textbf{0.934} \\
\bottomrule
\end{tabular*}
\end{table*}

\begin{figure*}[!t]
\centering
\setlength{\tabcolsep}{1pt}
\begin{tabular*}{\textwidth}{@{} c c c c c c c@{}}
\textbf{Inputs} & \textbf{pixelSplat} & \textbf{MVSplat} & \textbf{DepthSplat} & \textbf{NoPoSplat} & \textbf{Ours} &\textbf{GT RGB/Depth}\\ 
    \begin{minipage}[b]{0.075\textwidth}
        \includegraphics[width=\textwidth]{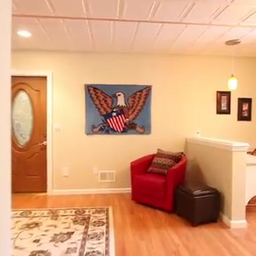}\\[3pt]
        \includegraphics[width=\textwidth]{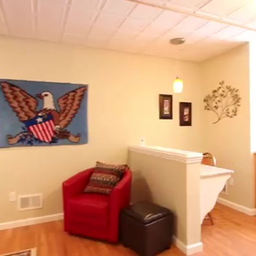}
    \end{minipage} 
    &
    \includegraphics[width=0.15\textwidth]{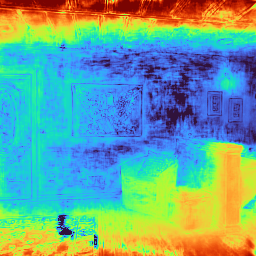}
    &
    \includegraphics[width=0.15\textwidth]{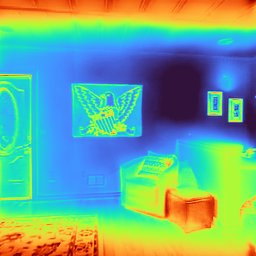}
    &
    \includegraphics[width=0.15\textwidth]{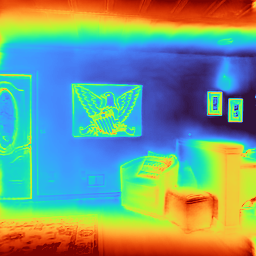}
    &
    \includegraphics[width=0.15\textwidth]{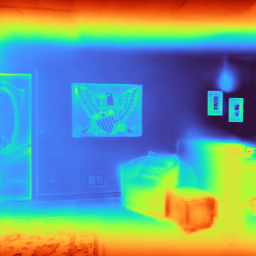}
    &
    \includegraphics[width=0.15\textwidth]{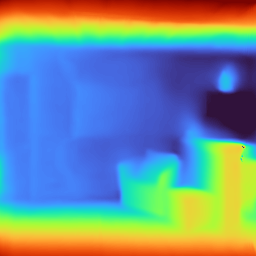}
    &
    \includegraphics[width=0.15\textwidth]{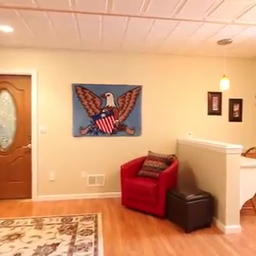}
    \\
    \begin{minipage}[b]{0.075\textwidth}
        \includegraphics[width=\textwidth]{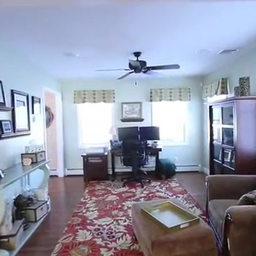}\\[3pt]
        \includegraphics[width=\textwidth]{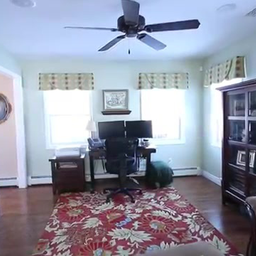}
    \end{minipage} 
    &
    \includegraphics[width=0.15\textwidth]{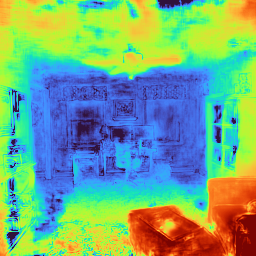}
    &
    \includegraphics[width=0.15\textwidth]{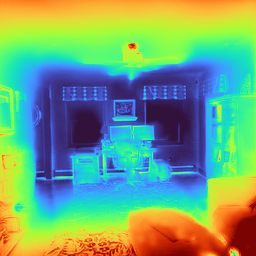}
    &
    \includegraphics[width=0.15\textwidth]{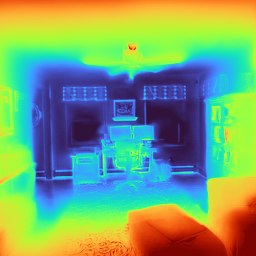}
    &
    \includegraphics[width=0.15\textwidth]{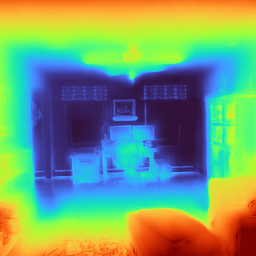}
    &
    \includegraphics[width=0.15\textwidth]{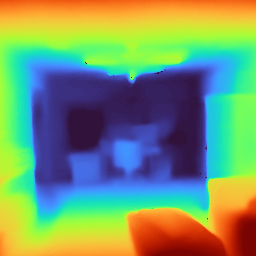}
    &
    \includegraphics[width=0.15\textwidth]{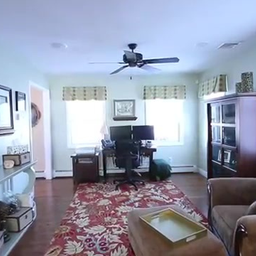}
    \\
    \begin{minipage}[b]{0.075\textwidth}
        \includegraphics[width=\textwidth]{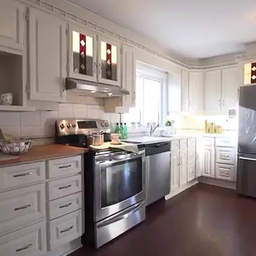}\\[3pt]
        \includegraphics[width=\textwidth]{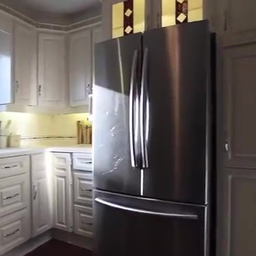}
    \end{minipage} 
    &
    \includegraphics[width=0.15\textwidth]{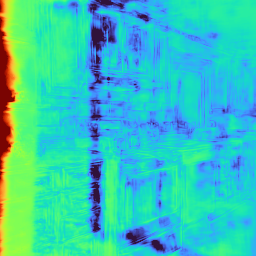}
    &
    \includegraphics[width=0.15\textwidth]{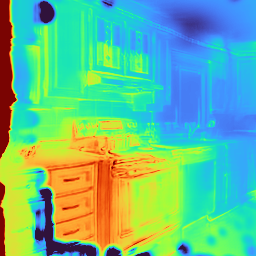}
    &
    \includegraphics[width=0.15\textwidth]{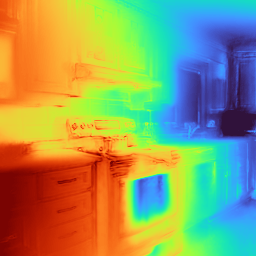}
    &
    \includegraphics[width=0.15\textwidth]{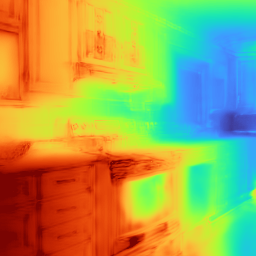}
    &
    \includegraphics[width=0.15\textwidth]{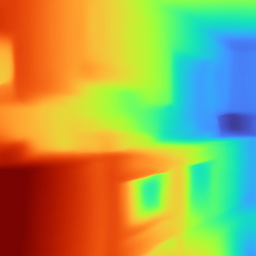}
    &
    \includegraphics[width=0.15\textwidth]{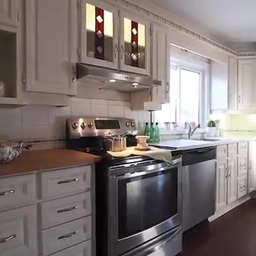}
    \\
    \begin{minipage}[b]{0.075\textwidth}
        \includegraphics[width=\textwidth]{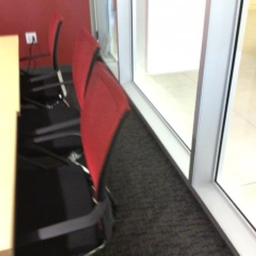}\\[3pt]
        \includegraphics[width=\textwidth]{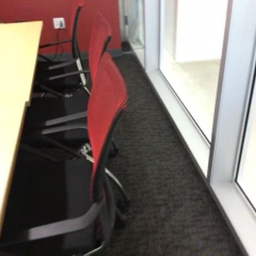}
    \end{minipage} 
    &
    \includegraphics[width=0.15\textwidth]{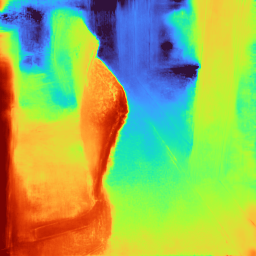}
    &
    \includegraphics[width=0.15\textwidth]{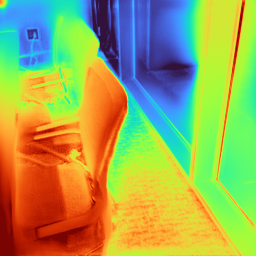}
    &
    \includegraphics[width=0.15\textwidth]{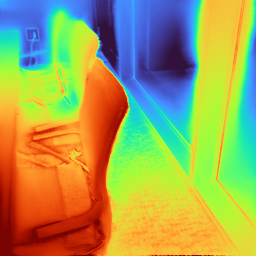}
    &
    \includegraphics[width=0.15\textwidth]{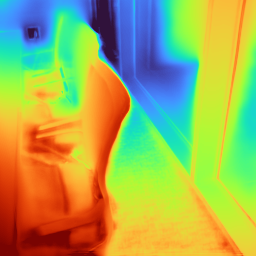}
    &
    \includegraphics[width=0.15\textwidth]{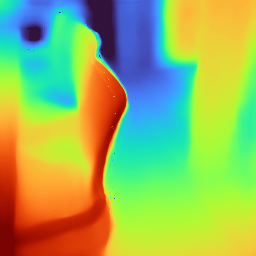}
    &
    \includegraphics[width=0.15\textwidth]{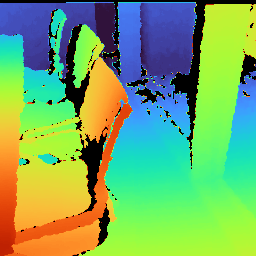}
    \\
    \begin{minipage}[b]{0.075\textwidth}
        \includegraphics[width=\textwidth]{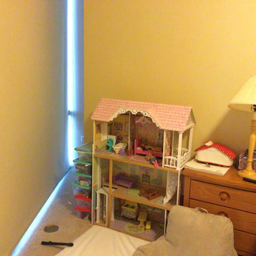}\\[3pt]
        \includegraphics[width=\textwidth]{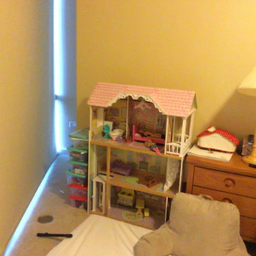}
    \end{minipage}
    &
    \includegraphics[width=0.15\textwidth]{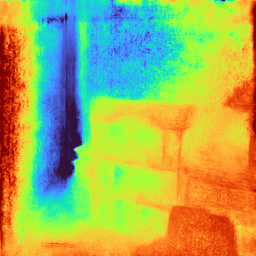}
    &
    \includegraphics[width=0.15\textwidth]{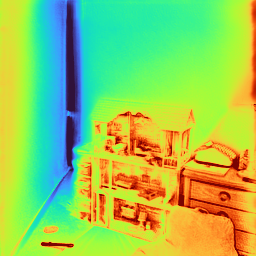}
    &
    \includegraphics[width=0.15\textwidth]{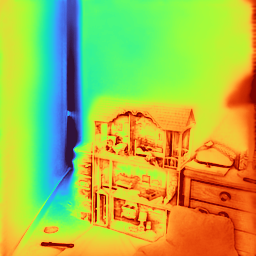}
    &
    \includegraphics[width=0.15\textwidth]{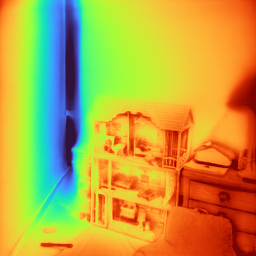}
    &
    \includegraphics[width=0.15\textwidth]{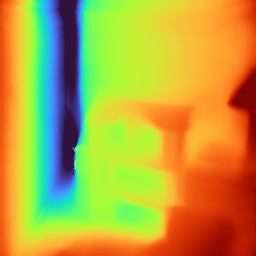}
    &
    \includegraphics[width=0.15\textwidth]{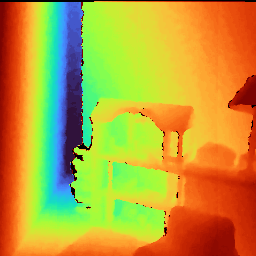}
    \\
    \begin{minipage}[b]{0.075\textwidth}
        \includegraphics[width=\textwidth]{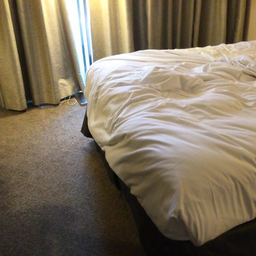}\\[3pt]
        \includegraphics[width=\textwidth]{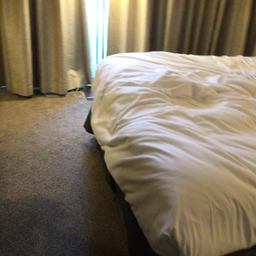}
    \end{minipage} 
    &
    \includegraphics[width=0.15\textwidth]{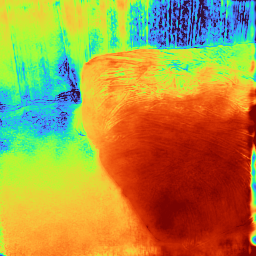}
    &
    \includegraphics[width=0.15\textwidth]{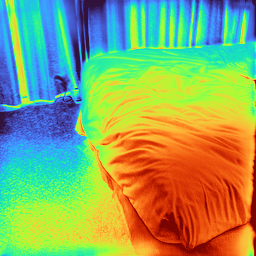}
    &
    \includegraphics[width=0.15\textwidth]{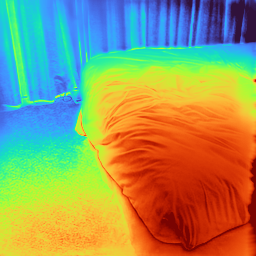}
    &
    \includegraphics[width=0.15\textwidth]{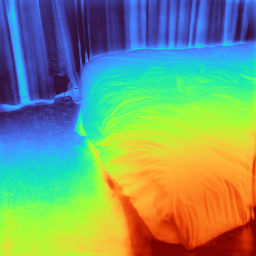}
    &
    \includegraphics[width=0.15\textwidth]{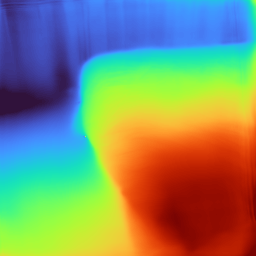}
    &
    \includegraphics[width=0.15\textwidth]{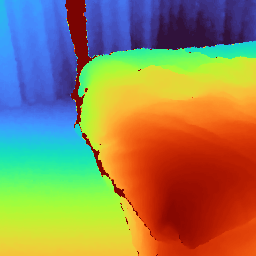}
    \\    
\end{tabular*}
\caption{\textbf{More qualitative comparison of novel-view rendered depth on RE10K~\cite{re10k} and ScanNet~\cite{scannet}}. pixelSplat depths are relatively geometrically consistent but noisy when the baseline is small. Large errors can be observed in the pixelSplat depths when the image overlap is small. Other baselines provide depth maps, which are hypersensitive to image texture. While some potentially meaningful fine structural edges are visible in these depth maps (see chair handles in row 4), the depth maps have many non-geometric ``fake edges'' (paintings in row 1, sofa in row 2, cabinet in row 3 -- to name a few). Our method provides geometrically consistent renderings in all these scenarios.}
\label{fig:supp_qualitative_nv_depth}
\end{figure*}

\subsection{Complete Novel-View Synthesis Comparisons}
\label{subsec:supp_complete_nvs}

For completeness, we report the full novel-view synthesis comparisons on RE10K~\cite{re10k}, ACID~\cite{acid}, and ScanNet~\cite{scannet}. These tables contextualize the ablation-focused synthesis results. They should be read as a check that the proposed priors preserve the view-synthesis behavior of generalizable Gaussian splatting while improving geometric consistency in the depth, mesh, and pose evaluations.

\begin{table*}[t]
    \centering
    \caption{
    \textbf{Complete novel-view synthesis comparison on RE10K~\cite{re10k}.}
    We report PSNR, SSIM, and LPIPS across the overlap regimes used in~\cite{noposplat}.
    These results provide context for the ablation-focused synthesis study; no target-pose optimization is used.
    }
    \begin{adjustbox}{max width=\textwidth}
    \setlength{\tabcolsep}{0.08cm}
    \renewcommand{\arraystretch}{1.1}
    \small
    \begin{tabular}{ll ccc ccc ccc ccc}
        \toprule
         & & \multicolumn{3}{c}{Small} & \multicolumn{3}{c}{Medium} & \multicolumn{3}{c}{Large} & \multicolumn{3}{c}{Average} \\
         \cmidrule(lr){3-5} \cmidrule(lr){6-8} \cmidrule(lr){9-11} \cmidrule(lr){12-14}
         & Method & PSNR$\uparrow$ & SSIM$\uparrow$ & LPIPS$\downarrow$ & PSNR$\uparrow$ & SSIM$\uparrow$ & LPIPS$\downarrow$ & PSNR$\uparrow$ & SSIM$\uparrow$ & LPIPS$\downarrow$ & PSNR$\uparrow$ & SSIM$\uparrow$ & LPIPS$\downarrow$ \\       
        \midrule
        \multirow{6}{*}{\shortstack[l]{\emph{Pose-} \\ \emph{required}}}
        & pixelNeRF~\cite{pixelnerf} 
        & 18.417 & 0.601 & 0.526 & 19.930 & 0.632 & 0.480 & 20.869 & 0.639 & 0.458 & 19.824 & 0.626 & 0.485 \\
        & AttnRend~\cite{du2023learning} 
        & 19.151 & 0.663 & 0.368 & 22.532 & 0.763 & 0.269 & 25.897 & 0.845 & 0.186 & 22.664 & 0.762 & 0.269 \\
        & pixelSplat~\cite{pixelsplat} 
         & 20.263         & 0.717           & 0.266  
         & 23.711         & 0.809           & 0.181  
         & 27.151         & 0.879           & 0.122  
         & 23.848         & 0.806           & 0.185  \\
        & MVSplat~\cite{mvsplat} 
         & 20.353   & 0.724       & 0.250  
         & 23.778   & 0.812       & 0.173  
         & 27.408   & 0.884       & 0.116  
         & 23.977   & 0.811       & 0.176  \\
        & FreeSplat~\cite{freesplat} 
         & 19.411   &  0.691   &  0.277
         & 22.839   &  0.790   &  0.192
         & 26.433   &  0.869   &  0.130 
         & 23.026   &  0.788   &  0.196 \\
        & DepthSplat~\cite{depthsplat} 
         & 22.820   & 0.798       & 0.193  
         & 25.383   & 0.851       & 0.145  
         & 28.317   & 0.900       & 0.104  
         & 25.595   & 0.852       & 0.145  \\
        \midrule
        \midrule
        \multirow{6}{*}{\shortstack[l]{\emph{Pose-} \\ \emph{free} 
        }}
        & Splatt3R~\cite{splatt3r} 
        & 14.352 & 0.475 & 0.472 & 15.529 & 0.502 & 0.425 & 15.817 & 0.483 & 0.421 & 15.318 & 0.490 & 0.436 \\
        & CoPoNeRF~\cite{coponerf} 
        & 17.393 & 0.585 & 0.462 & 18.813 & 0.616 & 0.392 & 20.464 & 0.652 & 0.318 & 18.938 & 0.619 & 0.388 \\
        & SelfSplat~\cite{selfsplat} 
         & 15.557  &  0.572  &  0.435  
         & 19.648  &  0.703  &  0.301 
         & 24.142  &  0.817  &  0.191 
         & 19.931  &  0.704  &  0.303    \\ 
        & NoPoSplat~\cite{noposplat} 
         & 21.097       & 0.723           & 0.237  
         & 23.191       & 0.779           & 0.187  
         & 25.107       & 0.817           & 0.144  
         & 23.244       & 0.778           & 0.187  \\
        & Ours \small{(3DGS)} 
         &  21.221  &  0.731   & 0.235         
         &  23.347  &  0.785   & 0.185         
         &  25.418  & 0.826    & 0.141 
         &  23.417  &  0.783   & 0.185         \\
        & Ours \small{(2DGS)} 
         & 21.377  & 0.739    &  0.234         
         & 23.426  & 0.787    & 0.184         
         & 25.459  & 0.827    & 0.141         
         & 23.504  & 0.787    & 0.184         \\
        \bottomrule
    \end{tabular}
    \end{adjustbox}
\label{tab:supp_re10k_nvs_full}
\end{table*}

\begin{table}[t]
    \centering
    \caption{\textbf{Zero-shot novel-view synthesis on ACID~\cite{acid} and ScanNet~\cite{scannet}.}
    All models are trained on RE10K~\cite{re10k} and evaluated without target-pose optimization.
    The table reports the pose-free comparison used to contextualize the ablation-focused synthesis analysis.}
    \begin{adjustbox}{max width=\columnwidth}
    \setlength{\tabcolsep}{0.08cm}
    \renewcommand{\arraystretch}{1.1}
    \small
    \begin{tabular}{l ccc ccc}
        \toprule
         & \multicolumn{3}{c}{ACID} & \multicolumn{3}{c}{ScanNet} \\
         \cmidrule(lr){2-4} \cmidrule(lr){5-7} 
         Method & PSNR$\uparrow$ & SSIM$\uparrow$ & LPIPS$\downarrow$ & PSNR$\uparrow$ & SSIM$\uparrow$ & LPIPS$\downarrow$ \\       
         \midrule
        NoPoSplat~\cite{noposplat} 
         &  23.379  &  0.684   &   0.237
         &  21.069  &  0.646   &   0.269     \\
        Ours \small{(3DGS)} 
         &  23.763  &  0.700   &  0.236 
         &  21.137  &  0.648   &  0.269     \\
        Ours \small{(2DGS)} 
         & 23.827  & 0.701   & 0.235   
         & 21.168  & 0.650   & 0.266  \\  
        \bottomrule
    \end{tabular}
    \end{adjustbox}
\label{tab:supp_acid_scannet_nvs_full}
\end{table}

The RE10K results in \Cref{tab:supp_re10k_nvs_full} follow the overlap regimes used in~\cite{noposplat}. The zero-shot results in \Cref{tab:supp_acid_scannet_nvs_full} use models trained only on RE10K and evaluated without target-pose optimization. Across these settings, the proposed priors yield small but consistent image-quality changes, while the geometry-centric evaluations show substantially larger gains.

\section{Architectures}
\label{sec:supp_architectures}
The proposed priors are architecture-agnostic. We therefore instantiate the same splat-prediction idea with two multi-view transformer backbones: a DUSt3R-style~\cite{dust3r} encoder (similar in spirit to~\cite{noposplat}) and a VGGT-style generalist geometry transformer~\cite{VGGT}. In both cases, the backbone produces per-image feature maps and multi-view aggregated features, and our Gaussian decoders predict per-pixel splat parameters (centers, scales, orientations, opacities, and colors). 

\noindent\textbf{DUSt3R-based variant.}
Our first instantiation builds on the pose-free, $N$-view transformer design pioneered by DUSt3R~\cite{dust3r}, which was also adopted in~\cite{noposplat} for generalizable Gaussian splatting, and adapts it to predict Gaussian splats with our 3DGS/2DGS parameterizations and priors. The architecture comprises three main components:
(i) a transformer-based image encoder,
(ii) two asymmetric multi-view feature aggregators, and
(iii) dense prediction decoders for Gaussian parameters.

The image encoder maps each RGB frame and its intrinsic parameters to a sequence of tokens (patch embeddings plus camera embeddings), which are processed independently by a ViT encoder. The resulting per-view features are then fused by two sets of cross-attention-based aggregators: one produces features for the reference (first) image, and the other produces features for the remaining images, conditioned on the reference. This asymmetric aggregation follows the spirit of DUSt3R, aligning all frames to a common canonical coordinate frame without requiring ground-truth poses.

Given the aggregated features, two DPT-style decoders predict Gaussian parameters. The first decoder regresses the Gaussian centers (i.e., 3D positions associated with each input pixel), while the second decoder predicts the remaining parameters (scales, orientations, opacities, and colors), optionally combining higher-resolution image features for appearance. Because all predicted Gaussians are expressed in a shared canonical frame, their union can be directly rendered from arbitrary viewpoints using the differentiable splatting pipeline, without explicit warping or known camera poses. Both the image tokenizer and the feature aggregators are built entirely from standard Vision Transformer blocks, without epipolar-specific attention or explicit multi-view cost volumes, keeping the architecture geometry-free and compatible with our priors.

\noindent\textbf{VGGT-based variant.}
For the VGGT-based instantiation~\cite{VGGT}, we retain the original pose and point-cloud branches and append a Gaussian-splat decoder that consumes the multi-view features to predict per-pixel Gaussian parameters. We do not use the VGGT depth branch to supervise the Gaussian representation. The decoder predicts centers, scales, orientations, opacities, and colors for 3DGS splats and is optimized through the rendered view-synthesis objective together with the proposed alignment and orientation priors. Because VGGT uses camera and point-map conventions that differ from the DUSt3R-style setup, we explicitly account for its intrinsic/extrinsic convention and include a pseudo-pose consistency term, following VGGT-based splatting practice~\cite{anysplat}, to stabilize the camera branch during training.

\section{Implementation Details}
\label{sec:supp_implementation_details}
\noindent\textbf{Common training setup.}
All models are trained in a generalizable splatting regime on the RealEstate10K (RE10K) training split~\cite{re10k}. We use two input images per sample and render three virtual novel views to minimize the view-synthesis loss. The same loss weights are used across all architectures and ablations: the alignment and orientation priors are weighted by $\lambda_a = 0.1$ and $\lambda_o = 0.05$. 

For the edge-aware weights in \Cref{eq:edge_aware_weights}, we set the robust scale to the $q{=}0.95$ quantile, i.e., $\eta=\mathrm{Quantile}_{0.95}(\{d_t^j\})$. We use fixed constants $(w_0,\kappa)=(10,4)$ and a small $\epsilon$ for numerical stability (we use $\epsilon=10^{-8}$ in all experiments). For the cosine-space penalty in \Cref{eq:orient_loss}, we use the Huber (SmoothL1) threshold $\delta=0.1$. We apply the scale regularization $\mathcal{L}_{\text{flat}}$ only for the 3DGS variant, with a fixed weight $\lambda_{\text{flat}}=1000$. All variants are evaluated on the same test splits and protocols described below. Source code and pretrained models are released on our project page to facilitate reproducibility.

\noindent\textbf{DUSt3R-based variant.}
For the DUSt3R-based backbone, training is performed on a cluster of 24 NVIDIA A100 (40\,GB) GPUs with a batch size of 6 per GPU (144 total), while all evaluations are run on a single NVIDIA A6000 GPU. We train for 18{,}751 iterations on RE10K using the setup above, with input images resized to $256{\times}256$.
The Gaussian decoders are optimized with a base learning rate of $2{\times}10^{-4}$, and the DUSt3R layers are updated with a reduced rate of $2{\times}10^{-5}$. In line with observations from~\cite{noposplat}, we found that training this architecture from scratch on RE10K is unstable in the fully self-supervised setting; instead, we initialize the backbone with MASt3R-pretrained weights~\cite{mast3r} and fine-tune it jointly with our Gaussian decoders. Competing DUSt3R-style baselines are allowed to use the same supervised backbone initialization for fairness. Under this setup, training the DUSt3R-based variant on RE10K takes approximately 6 hours.

\noindent\textbf{VGGT-based variant.}
For the VGGT-based backbone~\cite{VGGT}, we adopt the same dataset, number of source views, and number of rendered novel views as in the DUSt3R-based setup. During training, we cap the longer image side at 448 pixels and randomly vary the aspect ratio between 0.5 and 1.0. The Gaussian decoder attached to VGGT is optimized with a base learning rate of $2{\times}10^{-4}$, while the VGGT backbone is updated with a lower rate of $2{\times}10^{-5}$ to preserve its pretrained multi-view geometry prior. We initialize from VGGT pretrained weights and train the added Gaussian head end-to-end with the view-synthesis loss, the proposed geometric priors, and the pseudo-pose consistency term described above. We do not use pseudo-depth supervision for the Gaussian branch. The corresponding wall-clock training time for the VGGT-based variant is approximately $8$ hours on a single GPU with a batch size of 36.

\noindent\textbf{Evaluation protocols.}
To evaluate zero-shot generalization in pose estimation, geometry reconstruction, and novel-view synthesis, we use the same test sets and splits for all architectures and baselines. Specifically, we evaluate on the ACID split from~\cite{noposplat} and on the ScanNet test set~\cite{scannet}, which comprises 2000 indoor RGB-D image pairs. For ScanNet, novel views are obtained by uniformly sampling up to four intermediate viewpoints along the camera trajectory between each pair of source views used for pose evaluation, resulting in 1592 novel-view samples (out of 2000). These novel views are used consistently for depth, mesh, and novel-view synthesis evaluations.

\noindent\textbf{Baselines and retraining protocol.}
For all baselines, we use publicly released pretrained checkpoints whenever available. 
When multiple versions exist, we select models trained on RE10K~\cite{re10k} and at least at our input resolution of $256{\times}256$; if a model is trained on RE10K plus additional data or at higher resolution, we still use that checkpoint to give the baseline a slight advantage. 
If no RE10K-pretrained model is available (e.g., for FreeSplat~\cite{freesplat}), we retrain the method following the authors' original training protocol and hyperparameters to ensure a fair comparison.

For NoPoSplat~\cite{noposplat} specifically, we additionally retrain their model under our setup (RE10K dataset, iteration count, input resolution, and view-synthesis protocol). 
Minor deviations from the originally reported numbers are summarized in \Cref{tab:supp_noposplat_all_three}.

\begin{table*}[t]
  \centering
  \caption{\textbf{Comparison of our retrained NoPoSplat against the public checkpoint (NoPoSplat$^*$).}
    (a) Pose evaluation (with test-time photometric pose refinement) on RE10K~\cite{re10k}, ScanNet~\cite{scannet} and ACID~\cite{acid}. 
    (b) Depth estimation for novel views (with pose refinement) on ScanNet.  
    (c) Novel-view synthesis on RE10K.}
  \label{tab:supp_noposplat_all_three}
  \footnotesize\textbf{(a) Pose evaluation}\par
  \begin{adjustbox}{max width=\textwidth}
    \begin{tabular}{l ccc ccc ccc}
      \toprule
      & \multicolumn{3}{c}{RE10K}
      & \multicolumn{3}{c}{ScanNet-V1}
      & \multicolumn{3}{c}{ACID} \\
      \cmidrule(lr){2-4}\cmidrule(lr){5-7}\cmidrule(lr){8-10}
      Method & 5$^\circ\uparrow$ & 10$^\circ\uparrow$ & 20$^\circ\uparrow$
             & 5$^\circ\uparrow$ & 10$^\circ\uparrow$ & 20$^\circ\uparrow$
             & 5$^\circ\uparrow$ & 10$^\circ\uparrow$ & 20$^\circ\uparrow$ \\
      \midrule
      NoPoSplat$^*$ & 0.672 & 0.792 & 0.869 & 0.111 & 0.254 & 0.465 & 0.454 & 0.591 & 0.709 \\
      NoPoSplat     & 0.672 & 0.791 & 0.868 & 0.109 & 0.256 & 0.463 & 0.456 & 0.593 & 0.705 \\
      \bottomrule
    \end{tabular}
  \end{adjustbox}

  \vspace{1.2ex}  

  \footnotesize\textbf{(b) Depth estimation}\par
  \begin{adjustbox}{max width=\textwidth}
    \begin{tabular}{l ccc}
      \toprule
      & \multicolumn{3}{c}{Rendered Depth (Novel Views)} \\
      \cmidrule(lr){2-4}
      Method & Abs Rel$\downarrow$ & $\delta_1<1.10\uparrow$ & $\delta_1<1.25\uparrow$ \\
      \midrule
      NoPoSplat$^*$ & 0.127 & 0.564 & 0.859 \\
      NoPoSplat     & 0.126 & 0.567 & 0.861 \\
      \midrule
      Our baseline (No Prior 3DGS)  &  0.102 &  0.715  &  0.901  \\
      Our baseline (No Prior 2DGS)  &  0.114 &  0.692  &  0.884  \\
      \bottomrule
    \end{tabular}
  \end{adjustbox}

  \vspace{1.2ex}  

  \footnotesize\textbf{(c) Novel-view synthesis}\par
  \begin{adjustbox}{max width=\textwidth}
    \begin{tabular}{l ccc ccc ccc ccc}
      \toprule
       & \multicolumn{3}{c}{Small}
       & \multicolumn{3}{c}{Medium}
       & \multicolumn{3}{c}{Large}
       & \multicolumn{3}{c}{Average} \\
       \cmidrule(lr){2-4}\cmidrule(lr){5-7}\cmidrule(lr){8-10}\cmidrule(lr){11-13}
       Method 
         & PSNR$\uparrow$ & SSIM$\uparrow$ & LPIPS$\downarrow$
         & PSNR$\uparrow$ & SSIM$\uparrow$ & LPIPS$\downarrow$
         & PSNR$\uparrow$ & SSIM$\uparrow$ & LPIPS$\downarrow$
         & PSNR$\uparrow$ & SSIM$\uparrow$ & LPIPS$\downarrow$ \\
      \midrule
      NoPoSplat$^*$ 
       & 21.086 & 0.721 & 0.237  
       & 23.134 & 0.776 & 0.185  
       & 25.086 & 0.818 & 0.141  
       & 23.189 & 0.775 & 0.185  \\
      NoPoSplat 
       & 21.097 & 0.723 & 0.237  
       & 23.191 & 0.779 & 0.187  
       & 25.107 & 0.817 & 0.144  
       & 23.244 & 0.778 & 0.187  \\
      \bottomrule
    \end{tabular}
  \end{adjustbox}
\end{table*}

\section{Rendered Normal for 2DGS}
\label{sec:supp_rendered_normal}
In this subsection, we revisit the rendered normal–depth consistency loss introduced by 2DGS~\cite{2DGS} and compare it against our proposed orientation prior $\mathcal{L}_{\text{orient}}$. While $\mathcal{L}_{\text{orient}}$ is defined directly on Gaussian normals and can be applied to both 2DGS and 3DGS parameterizations, the 2DGS loss operates on \emph{rendered} surface normals and thus can only be instantiated and ablated for our 2DGS variant, where we explicitly render per-pixel normals.

For a 3D Gaussian corresponding to pixel $j$ in image $t$,
\[
\mathcal{G}_t^j=\bigl(\boldsymbol{\mu}_t^j, \alpha_t^j, \mathbf{\Sigma}_t^j, \boldsymbol{c}_t^j\bigr),
\]
3DGS~\cite{3dgs} first projects the mean and covariance to the image plane of a novel view. Let
\[
\mathbf{P}_f = \mathbf{K}_f \bigl[\mathbf{R}_f \mid \mathbf{T}_f\bigr] \in \mathbb{R}^{3\times4}
\]
be the projection matrix of view $f$. Dropping $(t,j,f)$ for clarity, the homogeneous image of the mean is
\[
\bar{\boldsymbol{\mu}} = \mathbf{P}\,[\boldsymbol{\mu}^\top \; 1]^{\!\top},
\qquad
\boldsymbol{\mu}' =
\begin{bmatrix}
\bar{\mu}_x / \bar{\mu}_z \\
\bar{\mu}_y / \bar{\mu}_z
\end{bmatrix}.
\]
Denoting by
\[
\mathbf{J} = \frac{\partial (\boldsymbol{\mu}'\,\bar{\mu}_z)}{\partial \boldsymbol{\mu}}
\]
the Jacobian of the local affine approximation of the perspective map, the (unnormalized) screen–space covariance is
\[
\mathbf{\Sigma}' = \mathbf{J}\,\mathbf{P}\,\mathbf{\Sigma}\,\mathbf{P}^{\!\top}\mathbf{J}^{\!\top},
\]
and we keep only its upper–left $2\times2$ block,
\(
\mathbf{\Sigma}'_{uv} = (\mathbf{\Sigma}')_{1:2,1:2}.
\)
The projected 2D Gaussian footprint is then
\[
\mathcal{G}'(u,v)
   = \exp\!\Bigl[
     -\tfrac12
     \bigl((u,v)^{\!\top}-\boldsymbol{\mu}'\bigr)^{\!\top}
     \mathbf{\Sigma}_{uv}^{\prime-1}
     \bigl((u,v)^{\!\top}-\boldsymbol{\mu}'\bigr)
     \Bigr].
\]

For a pixel $(u,v)$ in view $f$, novel-view RGB is rendered via front-to-back $\alpha$-blending of $K$ depth-sorted Gaussians:
\begin{align}
\hat{\mathbf{I}}_f(u,v)=
\sum_{k=1}^{K}
      \boldsymbol{c}_{k}\, w_k(u,v),
\quad \\
w_k(u,v)
= T_k(u,v)\,\alpha_{k}\,\mathcal{G}'_{k}(u,v),
\label{eq:supp_alpha_blend}
\end{align}
with transmittance
\[
T_k(u,v) = \prod_{j<k}\bigl(1-\alpha_{j}\,\mathcal{G}'_{j}(u,v)\bigr).
\]
The same weights $w_k(u,v)$ can be reused to render depth and surface normals for that view.

\paragraph{Rendered normal–depth consistency loss from 2DGS.}
2DGS~\cite{2DGS} proposes a \emph{rendered normal–depth consistency} loss that enforces agreement between rendered surface normals and normals estimated from the rendered depth map. Let $\mathbf{x} = (u,v)^\top$ and let $D_r(\mathbf{x})$ denote the rendered depth obtained by combining per-Gaussian depths $d_k(\mathbf{x})$ with the same weights as color:
\[
D_r(\mathbf{x}) = \sum_k w_k(\mathbf{x})\, d_k(\mathbf{x}).
\]
Let $\boldsymbol{n}_k$ be the (unit) normal associated with Gaussian $\mathcal{G}_k$. The rendered normal is then
\[
\boldsymbol{N}_r(\mathbf{x})
= \biggl\|
    \sum_{k} w_k(\mathbf{x})\,\boldsymbol{n}_k
  \biggr\|_{*},
\]
where $\|\cdot\|_{*}$ denotes vector normalization. A corresponding normal
$\widehat{\boldsymbol{N}}(\mathbf{x})$ can be estimated from $D_r(\mathbf{x})$ via finite differences and normalization. The rendered normal–depth consistency loss introduced in~\cite{2DGS} penalizes the angular discrepancy between these two normals:
\begin{equation}
\mathcal{L}_{\mathrm{RNC}}
  = \frac{1}{|\Omega|}
   \sum_{\mathbf{x}\in\Omega}
        \omega(\mathbf{x})\,
        \Bigl(1-\bigl\langle
                 \boldsymbol{N}_r(\mathbf{x}),
                 \widehat{\boldsymbol{N}}(\mathbf{x})
               \bigr\rangle\Bigr),
\label{eq:rendered_normal_consistency}
\end{equation}
where $\Omega$ is the set of valid pixels and
\[
\omega(\mathbf{x}) = \sum_k w_k(\mathbf{x})
\]
acts as an opacity-based confidence weight.

\paragraph{Comparison to our orientation prior.}
In our work, $\mathcal{L}_{\mathrm{RNC}}$ is not part of the final model; we implement it in the 2DGS variant purely as a baseline to compare against our proposed orientation prior $\mathcal{L}_{\text{orient}}$, which operates directly on Gaussian normals and is defined consistently for both 2DGS and 3DGS parameterizations. Conceptually, $\mathcal{L}_{\mathrm{RNC}}$ acts \emph{after} rasterization and thus enforces coherence between \emph{rendered} depth and normals, whereas $\mathcal{L}_{\text{orient}}$ provides direct supervision on the predicted Gaussian orientations, independent of the rasterizer.

\begin{figure*}[ht]
\centering
\begin{tabular*}{\textwidth}{c | c}
$\mathcal{L}_{\mathrm{RNC}}$ and $\mathcal{L}_{align}$ 
& $\mathcal{L}_\mathrm{RNC}$ Only\\ 
    \midrule
    \includegraphics[width=0.48\textwidth]{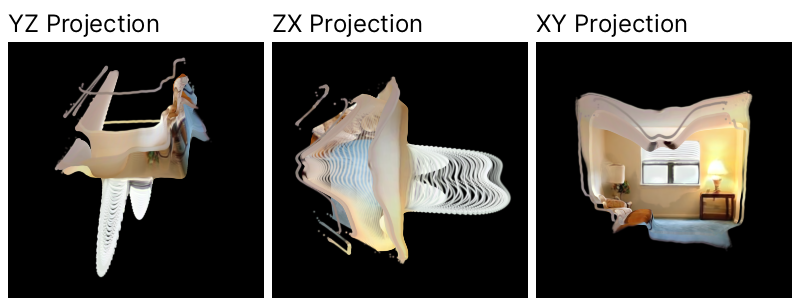}
    &
    \includegraphics[width=0.48\textwidth]{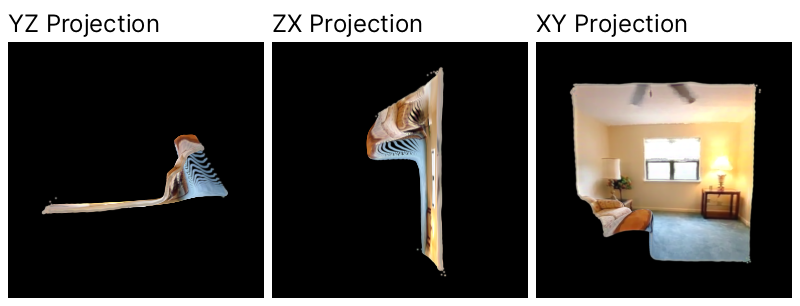}
\end{tabular*}
\caption{
\textbf{Failure modes of rendered normal–depth consistency compared to our orientation loss.}
We replace the proposed orientation prior $\mathcal{L}_{\text{orient}}$ with the rendered normal–depth consistency loss $\mathcal{L}_{\mathrm{RNC}}$~\cite{2DGS} and visualize the learned scene during training on RE10K~\cite{re10k}, by projecting the reconstructed 3D Gaussians onto three axis-aligned planes. 
Using $\mathcal{L}_{\mathrm{RNC}}$ without the alignment loss $\mathcal{L}_{\text{align}}$ yields severely degenerate reconstructions. 
Adding $\mathcal{L}_{\text{align}}$ recovers some structure but remains clearly inferior to the results obtained with $\mathcal{L}_{\text{align}} + \mathcal{L}_{\text{orient}}$ reported for the final model.}
\label{fig:supp_rendered_normal_loss_fail}
\end{figure*}

In practice, naively applying $\mathcal{L}_{\mathrm{RNC}}$ in our generalizable splatting setup often causes the optimization of Gaussian means and orientations to converge to a near-planar local minimum (see \Cref{fig:supp_rendered_normal_loss_fail}). Detaching the rendered depth from the computation graph alleviates this by treating the depth-derived normals as pseudo labels for Gaussian orientations, but this configuration still requires the alignment loss $\mathcal{L}_{\text{align}}$ to be effective. \Cref{tab:supp_normal_loss_comparison} compares models trained with $\mathcal{L}_{\text{align}} + \mathcal{L}_{\mathrm{RNC}}$ against our full model using $\mathcal{L}_{\text{align}} + \mathcal{L}_{\text{orient}}$, showing that our orientation prior is more stable and yields better geometry. Unless otherwise stated, \emph{all} 2DGS and 3DGS results reported in this manuscript use only $\mathcal{L}_{\text{align}}$ and $\mathcal{L}_{\text{orient}}$; $\mathcal{L}_{\mathrm{RNC}}$ is used solely as an ablation baseline in \Cref{tab:supp_normal_loss_comparison}.

\begin{table*}[t]
  \centering
    \caption{
    \textbf{Ablation of Gaussian orientation losses.}
    We compare a model trained with $\mathcal{L}_{\text{align}} + \mathcal{L}_{\mathrm{RNC}}$ against our full model trained with $\mathcal{L}_{\text{align}} + \mathcal{L}_{\text{orient}}$.
    (a) Pose evaluation with test-time refinement (same loss as training) on RE10K~\cite{re10k}, ScanNet~\cite{scannet}, and ACID~\cite{acid}. 
    (b) Novel-view depth estimation with pose refinement on ScanNet. 
    (c) Novel-view synthesis on RE10K.
    }
  \label{tab:supp_normal_loss_comparison}
  \footnotesize\textbf{(a) Pose evaluation}\par
  \begin{adjustbox}{max width=\textwidth}
    \begin{tabular}{l ccc ccc ccc}
      \toprule
      & \multicolumn{3}{c}{RE10K}
      & \multicolumn{3}{c}{ScanNet-V1}
      & \multicolumn{3}{c}{ACID} \\
      \cmidrule(lr){2-4}\cmidrule(lr){5-7}\cmidrule(lr){8-10}
      Method & 5$^\circ\uparrow$ & 10$^\circ\uparrow$ & 20$^\circ\uparrow$
             & 5$^\circ\uparrow$ & 10$^\circ\uparrow$ & 20$^\circ\uparrow$
             & 5$^\circ\uparrow$ & 10$^\circ\uparrow$ & 20$^\circ\uparrow$ \\
      \midrule
      Ours (2DGS+Align+RNC) 
      & 0.681 & 0.799 & 0.870 & 0.137 & 0.313 & 0.521 & 0.476 & 0.609 & 0.720 \\
      \best{Ours (2DGS+Align+Orient)}  & \best{0.686} & \best{0.802} & \best{0.875} & \best{0.153} & \best{0.334} & \best{0.541} & \best{0.478} & \best{0.609} & \best{0.723} \\
      \bottomrule
    \end{tabular}
  \end{adjustbox}

  \vspace{1.2ex}  

  \footnotesize\textbf{(b) Depth evaluation}\par
  \begin{adjustbox}{max width=\textwidth}
    \begin{tabular}{l ccc}
      \toprule
      & \multicolumn{3}{c}{Rendered Depth (Novel Views)} \\
      \cmidrule(lr){2-4}
      Method & Abs Rel$\downarrow$ & $\delta_1<1.10\uparrow$ & $\delta_1<1.25\uparrow$ \\
      \midrule
      Ours (2DGS+Align+RNC)      & 0.099 & 0.714 & 0.910 \\
      \best{Ours (2DGS+Align+Orient)}   & \best{0.082} & \best{0.743} & \best{0.931} \\
      \bottomrule
    \end{tabular}
  \end{adjustbox}

  \vspace{1.2ex}  

  \footnotesize\textbf{(c) Novel-view synthesis}\par
  \begin{adjustbox}{max width=\textwidth}
    \begin{tabular}{l ccc ccc ccc ccc}
      \toprule
       & \multicolumn{3}{c}{Small}
       & \multicolumn{3}{c}{Medium}
       & \multicolumn{3}{c}{Large}
       & \multicolumn{3}{c}{Average} \\
       \cmidrule(lr){2-4}\cmidrule(lr){5-7}\cmidrule(lr){8-10}\cmidrule(lr){11-13}
       Method 
         & PSNR$\uparrow$ & SSIM$\uparrow$ & LPIPS$\downarrow$
         & PSNR$\uparrow$ & SSIM$\uparrow$ & LPIPS$\downarrow$
         & PSNR$\uparrow$ & SSIM$\uparrow$ & LPIPS$\downarrow$
         & PSNR$\uparrow$ & SSIM$\uparrow$ & LPIPS$\downarrow$ \\
      \midrule
       Ours (2DGS+Align+RNC) 
       & 21.344 & 0.736 & 0.236  
       & 23.423 & 0.786 & 0.185  
       & 25.432 & 0.825 & 0.141  
       & 23.501 & 0.785 & 0.185  \\
       \best{Ours (2DGS+Align+Orient)} 
       & \best{21.377}  & \best{0.739}    & \best{0.234}      & \best{23.426}  & \best{0.787}    & \best{0.184}      & \best{25.459}  & \best{0.827}    & \best{0.141}      & \best{23.504}  & \best{0.787}    & \best{0.184}      \\       
      \bottomrule
    \end{tabular}
  \end{adjustbox}
\end{table*}

\section{Depth Rendering for Gaussian Splatting}
\label{sec:supp_depth_renderer}

To evaluate geometry (depth and meshes), we must convert the Gaussian representation into a per-pixel depth map. Along each camera ray, multiple Gaussians may contribute, so there is no uniquely defined ``depth.'' In this work we considered two choices, both consistent with standard alpha compositing.

Recall that for a pixel $(u,v)$ in view $f$, the rendered color is obtained by alpha-blending depth-sorted projected Gaussians $\mathcal{G}'_k$ as
\begin{align}
    \hat{\mathbf{I}}_f(u,v)
    &= \sum_{k=1}^K \boldsymbol{c}_k \, w_k(u,v),
    \label{eq:supp_synthesis_render} \\
    w_k(u,v)
    &= T_k(u,v)\,\alpha_k\,\mathcal{G}'_k(u,v), \\
    T_k(u,v)
    &= \prod_{i<k} \bigl(1 - \alpha_i\,\mathcal{G}'_i(u,v)\bigr),
\end{align}
where $\boldsymbol{c}_k$ and $\alpha_k$ denote the color and opacity of Gaussian $\mathcal{G}_k$, and $\mathcal{G}'_k(u,v)$ is its 2D footprint in the image plane of $\mathbf{I}_f$.

For depth rendering, we reuse the same weights $w_k(u,v)$ but replace the color $\boldsymbol{c}_k$ by the scalar depth $d_k(u,v)$ of Gaussian $\mathcal{G}_k$ in the camera coordinate system. For brevity, we write $\boldsymbol{x} = (u,v)$ and index the Gaussians along the ray by $i$.

\paragraph{Accumulated depth.}
A common practice is to treat depth as an additional ``channel'' and apply the same alpha-blending rule as for RGB. This yields an \emph{accumulated depth}
\begin{align}
    D_{\text{acc}}(\boldsymbol{x})
    = \sum_{i} w_i(\boldsymbol{x})\, d_i(\boldsymbol{x}),
    \label{eq:depth_acc}
\end{align}
where $w_i(\boldsymbol{x})$ is defined as in \eqref{eq:supp_synthesis_render}. This definition aggregates contributions from all Gaussians along the ray and implicitly couples depth with the overall opacity.

\paragraph{Expected depth.}
Alternatively, we can interpret the weights $w_i(\boldsymbol{x})$ as defining a discrete distribution along the ray and compute the \emph{expected depth}:
\begin{align}
    D_{\text{exp}}(\boldsymbol{x})
    = \frac{\sum_{i} w_i(\boldsymbol{x})\, d_i(\boldsymbol{x})}
           {\sum_{i} w_i(\boldsymbol{x})},
    \label{eq:depth_exp}
\end{align}
Compared to $D_{\text{acc}}$, this normalizes out the accumulated opacity and is less sensitive to residual transmittance or brightness variations.

\paragraph{Choice of baseline depth renderer.}
We implemented both $D_{\text{acc}}$ and $D_{\text{exp}}$ and compared them quantitatively for our model. In contrast to existing generalizable Gaussian splatting works that effectively rely on accumulated depth, we found that the expected depth $D_{\text{exp}}$ consistently yields lower depth errors and more stable TSDF fusion~\cite{tsdf-fusion}, resulting in higher-quality mesh reconstructions. The supplementary ablations report results both with and without the optional pose refinement described in \Cref{sec:supp_pose_refinement}, and all reported depth and mesh metrics use the expected-depth renderer unless explicitly stated otherwise.

Consequently, we adopt $D_{\text{exp}}$ in \eqref{eq:depth_exp} as our default depth rendering and use it for all ``no prior'' baselines. All proposed priors (alignment and orientation) are built on top of this expected-depth renderer and ablated accordingly.

\section{Test-time Pose Refinement}
\label{sec:supp_pose_refinement}
For two-view reconstruction, many different 3D configurations can explain the same image pair, so even if a model produces a self-consistent scene, its recovered camera poses may not coincide exactly with the ground-truth poses in the evaluation datasets. Following the spirit of prior pose-free works, including~\cite{noposplat}, we therefore allow an optional test-time pose refinement step for fair comparison against pose-aware baselines.

Given an input pair, we first run the network once to predict the Gaussian splats from the source views. During evaluation, these Gaussian parameters are \emph{frozen}, and we optimize the given camera pose by minimizing the same photometric objectives used at train time. Concretely, for a given view $f$ with ground-truth image $\mathbf{I}_f$ and rendered image $\hat{\mathbf{I}}_f(\mathbf{R}_f,\mathbf{T}_f)$, we solve a small gradient-based optimization problem
\[
\min_{\mathbf{R}_f,\mathbf{T}_f}
\;\mathcal{L}_{\text{synthesis}}(\mathbf{I}_f,\hat{\mathbf{I}}_f)
+ \lambda_a \mathcal{L}_{\text{align}}
+ \lambda_o \mathcal{L}_{\text{orient}},
\]
where only the pose parameters $(\mathbf{R}_f,\mathbf{T}_f)$ are updated and all Gaussian parameters remain fixed. For each model variant, we include exactly the same loss terms as used during training (e.g., $\mathcal{L}_{\text{align}}$ and/or $\mathcal{L}_{\text{orient}}$), so that the pose refinement is consistent with the learned priors.

We use this test-time pose refinement in two contexts:
(i) for pose evaluation, where the refined pose is compared against ground truth using rotation and translation error metrics; and
(ii) for depth evaluation, where we render depth maps from the refined camera pose to disentangle errors due to misaligned camera poses from errors in the underlying scene structure. All relevant pose and depth evaluations results are reported both \emph{with} and \emph{without} this pose refinement scheme, and whenever refinement is used it is explicitly stated in the corresponding table or figure caption.

\section{Mesh Reconstruction and Evaluation}
\label{sec:supp_mesh_evaluation_protocol}
\noindent\textbf{Predicted mesh reconstruction.}
For each ScanNet~\cite{scannet} test scene, we first predict a Gaussian-splatting representation from two source views using our model (either DUSt3R- or VGGT-based). The predicted Gaussians are then used to render depth maps from a virtual camera trajectory interpolated between the two source poses. Concretely, we sample a fixed number of intermediate viewpoints (20 in our implementation) by smoothly interpolating extrinsics between the two input cameras, and use our expected-depth renderer (Sec.~\Cref{sec:supp_depth_renderer}) to obtain per-view rendered depth maps along this path.
These depth maps, together with the corresponding camera intrinsics and extrinsics, are fused into a volumetric TSDF using TSDF-Fusion~\cite{tsdf-fusion}. We use a scene-adaptive voxel size (proportional to the scene radius) and a standard truncation distance (a small multiple of the voxel size), and extract a watertight surface mesh via Marching Cubes. Finally, we keep the largest connected component and remove small isolated clusters and degenerate faces, yielding the predicted mesh for that scene.

\medskip
\noindent\textbf{Ground-truth meshes and visibility cropping.}
ScanNet provides a metric-scale mesh for each scene. We rigidly transform this mesh into the coordinate frame where the first source camera is placed at the origin.

To avoid penalizing geometry that is never observed in the source views, we crop the ground-truth mesh to the region seen by the cameras. Specifically, we construct the union of viewing frusta of the source frames and intersect the mesh with this union. The resulting cropped mesh is used as the ground-truth surface for evaluation.

\medskip
\noindent\textbf{Global \texorpdfstring{Sim(3)}{Sim3} alignment.}
Because our pose-free model is trained without metric depth or absolute pose supervision, the recovered scene geometry and cameras are only defined up to a global Sim(3) transform (rotation, translation, and uniform scale). Before computing metrics, we therefore align each predicted mesh to its ground-truth counterpart using a single global Sim(3) transformation.
Concretely, we sample points from the surfaces of both the predicted mesh and the cropped ground-truth mesh, and run a point-to-point ICP procedure with scaling to estimate the best-fitting similarity transform between them. This transform is then applied to the predicted mesh, and all reconstruction metrics are computed in the resulting aligned, metric coordinate frame.

\medskip
\noindent\textbf{Evaluation metrics.}
Let $P = \{\mathbf{p}_i\}_{i=1}^{N_p}$ and $G = \{\mathbf{g}_j\}_{j=1}^{N_g}$ be point sets sampled uniformly from the aligned predicted mesh and the cropped ground-truth mesh, respectively. For each point $\mathbf{p}_i \in P$, we compute the distance to its nearest neighbor in $G$,
\[
d_{\text{pred}\to\text{gt}}(\mathbf{p}_i) = \min_{\mathbf{g}\in G} \|\mathbf{p}_i - \mathbf{g}\|_2,
\]
and similarly for each $\mathbf{g}_j \in G$ we compute
\[
d_{\text{gt}\to\text{pred}}(\mathbf{g}_j) = \min_{\mathbf{p}\in P} \|\mathbf{g}_j - \mathbf{p}\|_2.
\]

We report three standard reconstruction metrics~\cite{guo2022neural,occ_net}:

\begingroup
\setlength{\abovedisplayskip}{4pt}
\setlength{\belowdisplayskip}{4pt}
\setlength{\abovedisplayshortskip}{3pt}
\setlength{\belowdisplayshortskip}{3pt}
\begin{itemize}
    \item \textbf{Accuracy} (lower is better):
    \[
    \text{Acc} = \frac{1}{N_p} \sum_{i=1}^{N_p} d_{\text{pred}\to\text{gt}}(\mathbf{p}_i),
    \]
    measuring how close the predicted surface lies to the nearest ground-truth surface.

    \item \textbf{Completeness} (lower is better):
    \[
    \text{Comp} = \frac{1}{N_g} \sum_{j=1}^{N_g} d_{\text{gt}\to\text{pred}}(\mathbf{g}_j),
    \]
    measuring how well the predicted mesh covers the ground-truth surface.

    \item \textbf{Chamfer distance} (lower is better):
    \[
    \text{CD} = \frac{1}{2}\,\bigl(\text{Acc} + \text{Comp}\bigr),
    \]
    the symmetric Chamfer distance between $P$ and $G$.
\end{itemize}
\endgroup

Per-scene scores are averaged over all ScanNet test scenes where both predicted and ground-truth meshes are non-empty after reconstruction and cropping. These aggregated metrics are reported in \Cref{tab:mesh_reconstruction_scannet}.

%% file: references.bib
@String(CVPR= {IEEE Conf. Comput. Vis. Pattern Recog.})

@String(ICCV= {Int. Conf. Comput. Vis.})

@String(ECCV= {Eur. Conf. Comput. Vis.})

@String(TOG= {ACM Trans. Graph.})

@String(TIP  = {IEEE Trans. Image Process.})

@String(ACCV  = {ACCV})

@String(ICLR = {Int. Conf. Learn. Represent.})

@String(CVPR  = {CVPR})

@String(ICCV  = {ICCV})

@String(ECCV  = {ECCV})

@String(TOG   = {ACM TOG})

@String(TIP   = {IEEE TIP})

@String(ICLR  = {ICLR})

@inproceedings{dust3r,
  title={Dust3r: Geometric 3d vision made easy},
  author={Wang, Shuzhe and Leroy, Vincent and Cabon, Yohann and Chidlovskii, Boris and Revaud, Jerome},
  booktitle=CVPR,
  year={2024}
}

@article{mast3r,
  title={Grounding Image Matching in 3D with MASt3R},
  author={Leroy, Vincent and Cabon, Yohann and Revaud, J{\'e}r{\^o}me},
  journal={arXiv preprint arXiv:2406.09756},
  year={2024}
}

@inproceedings{pixelsplat,
  title={pixelsplat: 3d gaussian splats from image pairs for scalable generalizable 3d reconstruction},
  author={Charatan, David and Li, Sizhe Lester and Tagliasacchi, Andrea and Sitzmann, Vincent},
  booktitle=CVPR,
  year={2024}
}

@inproceedings{gps-gaussian,
  title={Gps-gaussian: Generalizable pixel-wise 3d gaussian splatting for real-time human novel view synthesis},
  author={Zheng, Shunyuan and Zhou, Boyao and Shao, Ruizhi and Liu, Boning and Zhang, Shengping and Nie, Liqiang and Liu, Yebin},
  booktitle=CVPR,
  year={2024}
}

@inproceedings{mvsplat,
  title={Mvsplat: Efficient 3d gaussian splatting from sparse multi-view images},
  author={Chen, Yuedong and Xu, Haofei and Zheng, Chuanxia and Zhuang, Bohan and Pollefeys, Marc and Geiger, Andreas and Cham, Tat-Jen and Cai, Jianfei},
  booktitle={ECCV},
  year={2024},
}

@inproceedings{depthsplat,
      title={DepthSplat: Connecting Gaussian Splatting and Depth},
      author={Xu, Haofei and Peng, Songyou and Wang, Fangjinhua and Blum, Hermann and Barath, Daniel and Geiger, Andreas and Pollefeys, Marc},
      booktitle={CVPR},
      year={2025}
    }

@inproceedings{pixelnerf,
  title={pixelnerf: Neural radiance fields from one or few images},
  author={Yu, Alex and Ye, Vickie and Tancik, Matthew and Kanazawa, Angjoo},
  booktitle=CVPR,
  year={2021}
}

@inproceedings{du2023learning,
  title={Learning to render novel views from wide-baseline stereo pairs},
  author={Du, Yilun and Smith, Cameron and Tewari, Ayush and Sitzmann, Vincent},
  booktitle=CVPR,
  year={2023}
}

@inproceedings{coponerf,
  title={Unifying Correspondence Pose and NeRF for Generalized Pose-Free Novel View Synthesis},
  author={Hong, Sunghwan and Jung, Jaewoo and Shin, Heeseong and Yang, Jiaolong and Kim, Seungryong and Luo, Chong},
  booktitle=CVPR,
  year={2024}
}

@inproceedings{garg2024direct,
  title={Direct Alignment for Robust NeRF Learning},
  author={Garg, Ravi and Chng, Shin-Fang and Lucey, Simon},
  booktitle=ACCV,
  year={2024},
}

@inproceedings{chng2024invert,
  title={Invertible Neural Warp for NeRF}, 
  author={Shin-Fang Chng and Ravi Garg and Hemanth Saratchandran and Simon Lucey},
  booktitle=ECCV,
  year={2024},
}

@article{splatt3r,
  title={Splatt3R: Zero-shot Gaussian Splatting from Uncalibrated Image Pairs},
  author={Smart, Brandon and Zheng, Chuanxia and Laina, Iro and Prisacariu, Victor Adrian},
  journal={arXiv preprint arXiv:2408.13912},
  year={2024}
}

@inproceedings{noposplat,
  title={No Pose, No Problem: Surprisingly Simple 3D Gaussian Splats from Sparse Unposed Images},
  author={Ye, Botao and Liu, Sifei and Xu, Haofei and Xueting, Li and Pollefeys, Marc and Yang, Ming-Hsuan and Songyou, Peng},
  booktitle=ICLR,
  year={2025}
}

@article{3dgs,
  title={3d gaussian splatting for real-time radiance field rendering},
  author={Kerbl, Bernhard and Kopanas, Georgios and Leimk{\"u}hler, Thomas and Drettakis, George},
  journal=TOG,
  year={2023},
}

@inproceedings{colmap,
  title={Structure-from-motion revisited},
  author={Schonberger, Johannes L and Frahm, Jan-Michael},
  booktitle=CVPR,
  year={2016}
}

@inproceedings{colmap-free3DGS,
  title={Colmap-free 3d gaussian splatting},
  author={Fu, Yang and Liu, Sifei and Kulkarni, Amey and Kautz, Jan and Efros, Alexei A and Wang, Xiaolong},
  booktitle=CVPR,
  year={2024}
}

@inproceedings{monogs,
  title={Gaussian splatting slam},
  author={Matsuki, Hidenobu and Murai, Riku and Kelly, Paul HJ and Davison, Andrew J},
  booktitle=CVPR,
  year={2024}
}

@article{nerfmm,
  title={NeRF--: Neural radiance fields without known camera parameters},
  author={Wang, Zirui and Wu, Shangzhe and Xie, Weidi and Chen, Min and Prisacariu, Victor Adrian},
  journal={arXiv preprint arXiv:2102.07064},
  year={2021}
}

@inproceedings{barf,
  title={Barf: Bundle-adjusting neural radiance fields},
  author={Lin, Chen-Hsuan and Ma, Wei-Chiu and Torralba, Antonio and Lucey, Simon},
  booktitle=ICCV,
  year={2021}
}

@inproceedings{niceslam,
  title={Nice-slam: Neural implicit scalable encoding for slam},
  author={Zhu, Zihan and Peng, Songyou and Larsson, Viktor and Xu, Weiwei and Bao, Hujun and Cui, Zhaopeng and Oswald, Martin R and Pollefeys, Marc},
  booktitle=CVPR,
  year={2022}
}

@inproceedings{nicerslam,
  title={Nicer-slam: Neural implicit scene encoding for rgb slam},
  author={Zhu, Zihan and Peng, Songyou and Larsson, Viktor and Cui, Zhaopeng and Oswald, Martin R and Geiger, Andreas and Pollefeys, Marc},
  booktitle={3DV},
  year={2024}
}

@inproceedings{wu2024_4dgs,
  title={4d gaussian splatting for real-time dynamic scene rendering},
  author={Wu, Guanjun and Yi, Taoran and Fang, Jiemin and Xie, Lingxi and Zhang, Xiaopeng and Wei, Wei and Liu, Wenyu and Tian, Qi and Wang, Xinggang},
  booktitle={CVPR},
  year={2024}
}

@inproceedings{yang2023real,
  title={Real-time photorealistic dynamic scene representation and rendering with 4d gaussian splatting},
  author={Yang, Zeyu and Yang, Hongye and Pan, Zijie and Zhang, Li},
  booktitle={ICLR},
  year={2024}
}

@inproceedings{yang_2024,
    title     = {Deformable 3D Gaussians for High-Fidelity Monocular Dynamic Scene Reconstruction},
    author    = {Yang, Ziyi and Gao, Xinyu and Zhou, Wen and Jiao, Shaohui and Zhang, Yuqing and Jin, Xiaogang},
    booktitle = {CVPR},
    year      = {2024},
}

@inproceedings{li_2024,
    title     = {Spacetime Gaussian Feature Splatting for Real-Time Dynamic View Synthesis},
    author    = {Li, Zhan and Chen, Zhang and Li, Zhong and Xu, Yi},
    booktitle = {CVPR},
    year      = {2024},
}

@inproceedings{cut3r,
  title = {Continuous 3D Perception Model with Persistent State},
  author = {Qianqian Wang and Yifei Zhang and Aleksander Holynski and Alexei A. Efros and Angjoo Kanazawa},
  booktitle={CVPR},
  year = {2025},
}

@inproceedings{splatam,
    title={SplaTAM: Splat, Track \& Map 3D Gaussians for Dense RGB-D SLAM},
    author={Keetha, Nikhil and Karhade, Jay and Jatavallabhula, Krishna Murthy and Yang, Gengshan and Scherer, Sebastian and Ramanan, Deva and Luiten, Jonathon},
    booktitle={CVPR},
    year={2024}
}

@inproceedings{zhu2024fsgs,
  title={Fsgs: Real-time few-shot view synthesis using gaussian splatting},
  author={Zhu, Zehao and Fan, Zhiwen and Jiang, Yifan and Wang, Zhangyang},
  booktitle={ECCV},
  year={2024},
}

@inproceedings{li2024dngaussian,
    title={DNGaussian: Optimizing Sparse-View 3D Gaussian Radiance Fields with Global-Local Depth Normalization}, 
    author={Jiahe Li and Jiawei Zhang and Xiao Bai and Jin Zheng and Xin Ning and Jun Zhou and Lin Gu},
    booktitle={CVPR},
    year={2024}
}

@inproceedings{xiong2023sparsegs,
title = {SparseGS: Real-Time 360° Sparse View Synthesis using Gaussian Splatting},
author = {Xiong, Haolin and Muttukuru, Sairisheek and Upadhyay, Rishi and Chari, Pradyumna and Kadambi, Achuta},
booktitle = {3DV},
year = {2025},
}

@article{freesplat,
  title={Freesplat: Generalizable 3d gaussian splatting towards free view synthesis of indoor scenes},
  author={Wang, Yunsong and Huang, Tianxin and Chen, Hanlin and Lee, Gim Hee},
  journal={NeurIPS},
  year={2024}
}

@article{efreesplat,
  title={Epipolar-Free 3D Gaussian Splatting for Generalizable Novel View Synthesis},
  author={Min, Zhiyuan and Luo, Yawei and Sun, Jianwen and Yang, Yi},
  journal={NeurIPS},
  year={2024}
}

@article{binocular3dgs,
  title={Binocular-guided 3d gaussian splatting with view consistency for sparse view synthesis},
  author={Han, Liang and Zhou, Junsheng and Liu, Yu-Shen and Han, Zhizhong},
  journal={NeurIPS},
  year={2024}
}

@article{hisplat,
  title={Hisplat: Hierarchical 3d gaussian splatting for generalizable sparse-view reconstruction},
  author={Tang, Shengji and Ye, Weicai and Ye, Peng and Lin, Weihao and Zhou, Yang and Chen, Tao and Ouyang, Wanli},
  journal={ICLR},
  year={2025}
}

@inproceedings{s2gaussian,
  title={S2Gaussian: Sparse-View Super-Resolution 3D Gaussian Splatting},
  author={Wan, Ziyu and Gao, Hao and Xiong, Rui and Du, Fang},
  booktitle={CVPR},
  year={2025}
}

@article{selfsplat,
  title={SelfSplat: Pose-Free and 3D-Prior-Free Generalizable 3D Gaussian Splatting},
  author={Kang, Seunghyun and Lee, Hyunwoo and Chae, Hyeongju},
  journal={CVPR},
  year={2025}
}

@inproceedings{li2024sgs,
  title={Sgs-slam: Semantic gaussian splatting for neural dense slam},
  author={Li, Mingrui and Liu, Shuhong and Zhou, Heng and Zhu, Guohao and Cheng, Na and Deng, Tianchen and Wang, Hongyu},
  booktitle={ECCV},
  year={2024},
}

@inproceedings{chung2024depth,
  title={Depth-regularized optimization for 3d gaussian splatting in few-shot images},
  author={Chung, Jaeyoung and Oh, Jeongtaek and Lee, Kyoung Mu},
  booktitle={CVPR},
  year={2024}
}

@inproceedings{2DGS,
  title={2d gaussian splatting for geometrically accurate radiance fields},
  author={Huang, Binbin and Yu, Zehao and Chen, Anpei and Geiger, Andreas and Gao, Shenghua},
  booktitle={ACM SIGGRAPH 2024},
  year={2024}
}

@article{sugar,
title={SuGaR: Surface-Aligned Gaussian Splatting for Efficient 3D Mesh Reconstruction and High-Quality Mesh Rendering},
author={Gu{\'e}don, Antoine and Lepetit, Vincent},
journal={CVPR},
year={2024}
}

@inproceedings{roma1,
  title={RoMa: Robust dense feature matching},
  author={Edstedt, Johan and Sun, Qiyu and B{\"o}kman, Georg and Wadenb{\"a}ck, M{\aa}rten and Felsberg, Michael},
  booktitle=CVPR,
  year={2024}
}

@article{re10k,
  author = {Zhou, Tinghui and Tucker, Richard and Flynn, John and Fyffe, Graham and Snavely, Noah},
  title = {Stereo magnification: learning view synthesis using multiplane images},
  year = {2018},
  journal = TOG,
}

@inproceedings{acid,
  title={Infinite nature: Perpetual view generation of natural scenes from a single image},
  author={Liu, Andrew and Tucker, Richard and Jampani, Varun and Makadia, Ameesh and Snavely, Noah and Kanazawa, Angjoo},
  booktitle=ICCV,
  year={2021}
}

@inproceedings{scannet,
  title={Scannet: Richly-annotated 3d reconstructions of indoor scenes},
  author={Dai, Angela and Chang, Angel X and Savva, Manolis and Halber, Maciej and Funkhouser, Thomas and Nie{\ss}ner, Matthias},
  booktitle=CVPR,
  year={2017}
}

@inproceedings{lpips,
  title={The unreasonable effectiveness of deep features as a perceptual metric},
  author={Zhang, Richard and Isola, Phillip and Efros, Alexei A and Shechtman, Eli and Wang, Oliver},
  booktitle=CVPR,
  year={2018}
}

@article{ssim,
  title={Image quality assessment: from error visibility to structural similarity},
  author={Wang, Zhou and Bovik, Alan C and Sheikh, Hamid R and Simoncelli, Eero P},
  journal=TIP,
  year={2004},
}

@inproceedings{dpt,
  title={Vision transformers for dense prediction},
  author={Ranftl, Ren{\'e} and Bochkovskiy, Alexey and Koltun, Vladlen},
  booktitle=ICCV,
  year={2021}
}

@article{tnt,
  title={Tanks and temples: Benchmarking large-scale scene reconstruction},
  author={Knapitsch, Arno and Park, Jaesik and Zhou, Qian-Yi and Koltun, Vladlen},
  journal=TOG,
  year={2017},
}

@inproceedings{garg2016,
  title={Unsupervised cnn for single view depth estimation: Geometry to the rescue},
  author={Garg, Ravi and Bg, Vijay Kumar and Carneiro, Gustavo and Reid, Ian},
  booktitle={ECCV},
  year={2016},
}

@inproceedings{godard2017,
  title={Unsupervised monocular depth estimation with left-right consistency},
  author={Godard, Cl{\'e}ment and Mac Aodha, Oisin and Brostow, Gabriel J},
  booktitle={CVPR},
  year={2017}
}

@inproceedings{zhou2017,
  title={Unsupervised learning of depth and ego-motion from video},
  author={Zhou, Tinghui and Brown, Matthew and Snavely, Noah and Lowe, David G},
  booktitle={CVPR},
  year={2017}
}

@inproceedings{zhan2018,
  title={Unsupervised learning of monocular depth estimation and visual odometry with deep feature reconstruction},
  author={Zhan, Huangying and Garg, Ravi and Weerasekera, Chamara Saroj and Li, Kejie and Agarwal, Harsh and Reid, Ian},
  booktitle={CVPR},
  year={2018}
}

@InProceedings{godard2019,
author = {Godard, Clement and Mac Aodha, Oisin and Firman, Michael and Brostow, Gabriel J.},
title = {Digging Into Self-Supervised Monocular Depth Estimation},
booktitle = {ICCV},
year = {2019}
}

@inproceedings{mvsnet,
  title={MVSNet: Depth Inference for Unstructured Multi-View Stereo},
  author={Yao, Yao and Luo, Zixin and Li, Shiwei and Fang, Tian and Quan, Long},
  booktitle={ECCV},
  year={2018}
}

@inproceedings{casmvsnet,
  title={Cascade cost volume for high-resolution multi-view stereo and stereo matching},
  author={Gu, Xiaodong and Fan, Zhiwen and Zhu, Siyu and Dai, Zuozhuo and Tan, Feitong and Tan, Ping},
  booktitle={CVPR},
  year={2020}
}

@inproceedings{patchmatchnet,
  title={Patchmatchnet: Learned multi-view patchmatch stereo},
  author={Wang, Fangjinhua and Galliani, Silvano and Vogel, Christoph and Speciale, Pablo and Pollefeys, Marc},
  booktitle={CVPR},
  year={2021}
}

@inproceedings{eigen2014depth, title={Depth Map Prediction from a Single Image using a Multi-Scale Deep Network}, author={Eigen, David and Puhrsch, Christian and Fergus, Rob}, booktitle={NeurIPS}, year={2014}}

@inproceedings{laina2016deeper, title={Deeper Depth Prediction with Fully Convolutional Residual Networks}, author={Laina, Iro and Rupprecht, Christian and Belagiannis, Vasileios and Tombari, Federico and Navab, Nassir}, booktitle={3DV}, year={2016}}

@inproceedings{fu2018deep, title={Deep Ordinal Regression Network for Monocular Depth Estimation}, author={Fu, Huan and Gong, Mingming and Wang, Chaohui and Batmanghelich, Kayhan and Tao, Dacheng}, booktitle={CVPR}, year={2018}}

@article{lee2019bts, title={BTS: Depth Estimation via Local Planar Guidance}, author={Lee, Jin Han and Bae, Youngbok and Han, In So Kweon}, journal={arXiv preprint arXiv:1907.10326}, year={2019}}

@inproceedings{chang2018pyramid, title={Pyramid Stereo Matching Network}, author={Chang, Jia-Ren and Chen, Yong-Sheng}, booktitle={CVPR}, year={2018}}

@inproceedings{huang2018deepmvs, title={DeepMVS: Learning Multi-view Stereopsis}, author={Huang, Jia-Bin and Matthews, Iain and Kienzle, Wolf}, booktitle={CVPR}, year={2018}}

@inproceedings{ummenhofer2017demon, title={DeMoN: Depth and Motion Network for Learning Monocular Stereo}, author={Ummenhofer, Benjamin and Zhou, Hao and Uhrig, Jonas and Mayer, Nikolaus and Ilg, Eddy and Dosovitskiy, Alexey and Brox, Thomas}, booktitle={CVPR}, year={2017}}

@article{unimatch,
  title={Unifying flow, stereo and depth estimation},
  author={Xu, Haofei and Zhang, Jing and Cai, Jianfei and Rezatofighi, Hamid and Yu, Fisher and Tao, Dacheng and Geiger, Andreas},
  journal={IEEE Transactions on Pattern Analysis and Machine Intelligence},
  volume={45},
  number={11},
  pages={13941--13958},
  year={2023},
  publisher={IEEE}
}

@inproceedings{eigen2015predicting,
  title={Predicting depth, surface normals and semantic labels with a common multi-scale convolutional architecture},
  author={Eigen, David and Fergus, Rob},
  booktitle={Proceedings of the IEEE international conference on computer vision},
  pages={2650--2658},
  year={2015}
}

@inproceedings{tsdf-fusion,
  title={A volumetric method for building complex models from range images},
  author={Curless, Brian and Levoy, Marc},
  booktitle={Proceedings of the 23rd annual conference on Computer graphics and interactive techniques},
  pages={303--312},
  year={1996}
}

@inproceedings{guo2022neural,
  title={Neural 3d scene reconstruction with the manhattan-world assumption},
  author={Guo, Haoyu and Peng, Sida and Lin, Haotong and Wang, Qianqian and Zhang, Guofeng and Bao, Hujun and Zhou, Xiaowei},
  booktitle={Proceedings of the IEEE/CVF conference on computer vision and pattern recognition},
  pages={5511--5520},
  year={2022}
}

@article{anysplat,
  title={AnySplat: Feed-forward 3D Gaussian Splatting from Unconstrained Views},
  author={Jiang, Lihan and Mao, Yucheng and Xu, Linning and Lu, Tao and Ren, Kerui and Jin, Yichen and Xu, Xudong and Yu, Mulin and Pang, Jiangmiao and Zhao, Feng and others},
  journal={arXiv preprint arXiv:2505.23716},
  year={2025}
}

@inproceedings{flash3d,
  title={Flash3d: Feed-forward generalisable 3d scene reconstruction from a single image},
  author={Szymanowicz, Stanislaw and Insafutdinov, Eldar and Zheng, Chuanxia and Campbell, Dylan and Henriques, Joao F and Rupprecht, Christian and Vedaldi, Andrea},
  booktitle={2025 International Conference on 3D Vision (3DV)},
  pages={670--681},
  year={2025},
  organization={IEEE},
}

@inproceedings{VGGT,
  title={VGGT: Visual Geometry Grounded Transformer},
  author={Wang, Jianyuan and Chen, Minghao and Karaev, Nikita and Vedaldi, Andrea and Rupprecht, Christian and Novotny, David},
  booktitle={Proceedings of the IEEE/CVF Conference on Computer Vision and Pattern Recognition},
  year={2025},
}

@inproceedings{occ_net,
  title={Occupancy networks: Learning 3d reconstruction in function space},
  author={Mescheder, Lars and Oechsle, Michael and Niemeyer, Michael and Nowozin, Sebastian and Geiger, Andreas},
  booktitle={Proceedings of the IEEE/CVF conference on computer vision and pattern recognition},
  pages={4460--4470},
  year={2019}
}

@misc{flare,
      title={FLARE: Feed-forward Geometry, Appearance and Camera Estimation from Uncalibrated Sparse Views}, 
      author={Shangzhan Zhang and Jianyuan Wang and Yinghao Xu and Nan Xue and Christian Rupprecht and Xiaowei Zhou and Yujun Shen and Gordon Wetzstein},
      year={2025},
      eprint={2502.12138},
      archivePrefix={arXiv},
      primaryClass={cs.CV},
      url={https://arxiv.org/abs/2502.12138}, 
}
